\documentclass[10pt,twocolumn,letterpaper]{article}
\usepackage{cvpr}      
\usepackage{graphicx}
\usepackage{amsmath}
\usepackage{amssymb}
\usepackage{makecell}
\usepackage{booktabs}
\usepackage{cuted}

\usepackage[pagebackref,breaklinks,colorlinks]{hyperref}
\usepackage[capitalize]{cleveref}
\newcommand{\para}[1]{\noindent\textbf{#1}.}

\crefname{section}{Sec.}{Secs.}
\Crefname{section}{Section}{Sections}
\Crefname{table}{Table}{Tables}
\crefname{table}{Tab.}{Tabs.}

\def\cvprPaperID{4969} 
\def\confName{CVPR}
\def\confYear{2022}

\begin{document}

\title{ImFace: A Nonlinear 3D Morphable Face Model \\
with Implicit Neural Representations}

\author{
Mingwu Zheng\textsuperscript{\rm 1,2},
Hongyu Yang\textsuperscript{\rm 3},
Di Huang\textsuperscript{\rm 1,2}\thanks{Corresponding author.},
Liming Chen\textsuperscript{\rm 4}\\
\textsuperscript{\rm 1} State Key Laboratory of Software Development Environment, Beihang University, China\\
\textsuperscript{\rm 2} School of Computer Science and Engineering, Beihang University, China\\
\textsuperscript{\rm 3} Institute of Artificial Intelligence, Beihang University, China\\
\textsuperscript{\rm 4} LIRIS, \'Ecole Centrale de Lyon, France\\
{\tt\small \{zhengmingwu,hongyuyang,dhuang\}@buaa.edu.cn, liming.chen@ec-lyon.fr}
}

\maketitle

\begin{abstract}
  Precise representations of 3D faces are beneficial to various computer vision and graphics applications. Due to the data discretization and model linearity, however, it remains challenging to capture accurate identity and expression clues in current studies. This paper presents a novel 3D morphable face model, namely ImFace, to learn a nonlinear and continuous space with implicit neural representations. It builds two explicitly disentangled deformation fields to model complex shapes associated with identities and expressions, respectively, and designs an improved learning strategy to extend embeddings of expressions to allow more diverse changes. We further introduce a Neural Blend-Field to learn sophisticated details by adaptively blending a series of local fields. In addition to ImFace, an effective preprocessing pipeline is proposed to address the issue of watertight input requirement in implicit representations, enabling them to work with common facial surfaces for the first time. Extensive experiments are performed to demonstrate the superiority of ImFace. Our code is available at \url{https://github.com/MingwuZheng/ImFace}.
\end{abstract}

\vspace{-5mm}
\section{Introduction}
\vspace{-1mm}
\label{sec:intro}

3D Morphable Face Models (3DMMs) are well-reputed statistical models, established by learning techniques upon prior distributions of facial shapes and textures from a set of samples with dense correspondence, aiming at rendering realistic faces of a high variety. Since a morphable representation is unique across different downstream tasks where the geometry and appearance are separately controllable, 3DMMs are pervasively exploited in many face analysis applications in the field of computer vision, computer graphics, biometrics, and medical imaging~\cite{aldrian2012inverse, blanz2003face, hu2016face, staal2015describing}.

In 3DMMs, the most fundamental issue lies in the way to generate latent morphable representations, and during the past two decades, along with data improvement in scale, diversity and quality~\cite{cao2013facewarehouse, booth2018large, li2017flame, yang2020facescape}, remarkable progresses have been achieved. The methods are initially linear model based~\cite{blanz1999morphable, patel20093d, bfm09} and further extended to multilinear model based~\cite{vlasic2006face, brunton2014multilinear, bolkart2015groupwise}, where different modes are individually encoded. Unfortunately, for the relatively limited representation ability of linear models, these methods are not so competent at handling the cases with complicated variations, \eg exaggerated expressions. In the context of deep learning, a number of nonlinear models have been investigated with the input of 2D images~\cite{tran2018nonlinear, tran2019towards} or 3D meshes~\cite{bagautdinov2018modeling, ranjan2018generating, bouritsas2019neural, chen2021learning, cheng2019arxiv} by using Convolutional Neural Networks or Graph Neural Networks. They indeed deliver performance gains; however, restricted by the resolution of discrete representing strategies on input data, facial priors are not sufficiently captured, incurring loss of shape details. Besides, all current methods are dependent on the preposed procedure of point-to-point correspondence~\cite{gilani2017dense, abrevaya2018spatiotemporal, liu20193d, bahri2021shape}, but face registration itself remains challenging.

\begin{figure}
  \centering
   \setlength{\abovecaptionskip}{5pt}
   \setlength{\belowcaptionskip}{0pt}
   \includegraphics[width=1\linewidth]{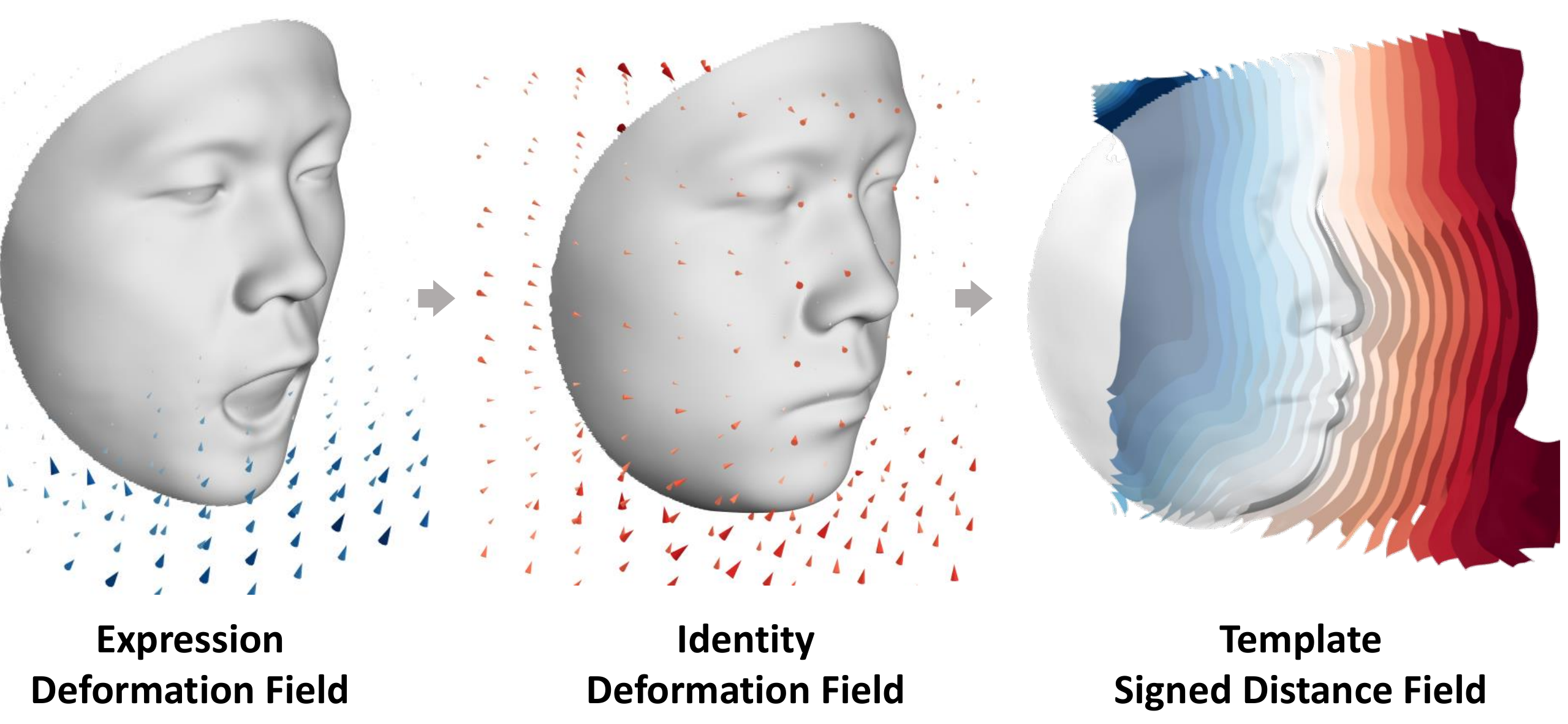}
    \vspace{-5mm}
   \caption{ImFace encodes complex face variations by two explicitly disentangled deformation fields with respect to a template face, resulting in a morphable implicit representation for 3D faces.}
   \label{fig:sample}
   \vspace{-5mm}
\end{figure}

Recently, several studies on Implicit Neural Representations (INRs) \cite{park2019deepsdf, mescheder2019occupancy, chen2019learning, icml2020_2086, pmlr-v139-lipman21a} have shown that 3D geometries can be precisely modeled by learning continuous deep implicit functions. They describe an input observation as a low-dimensional shape embedding and estimate the Signed Distance Function (SDF) or the occupancy value of a query point so that the surface of an arbitrary resolution and topology can be defined by an isocontour. Due to the continuous parameterization and consistent representation, INRs prove superior to the discrete voxels, pointclouds and meshes, and report decent results in shape reconstruction \cite{xu2019disn, wang2021neus, genova2019iccv,zhang2021learning} and surface registration \cite{deng2021deformed, liu2020nips, zheng2021deep}. Such an advantage suggests an alternative to 3DMM that can fulfill accurate correspondence and fine-grained modeling in a unified network. Nevertheless, unlike the objects with apparent shape differences and limited non-rigid variations such as indoor scenes and human bodies, all facial surfaces look very similar but include more complex deformations, where multiple identities and rich expressions deeply interweave with each other, making current INR methods problematic in face modeling, as evidenced by the preliminary attempt~\cite{yenamandra2021i3dmm}. Another difficulty is that implicit functions primarily require watertight input, which is not friendly to facial surfaces. 

This paper proposes a novel 3D face morphable model, namely ImFace, which substantially upgrades conventional 3DMMs by learning INRs. To capture nonlinear facial geometry changes, ImFace builds separate INR sub-networks to explicitly disentangle shape morphs into two deformation fields for identity and expression respectively (as Fig.~\ref{fig:sample} shows), and an improved embedding learning strategy is introduced to extend the latent space of expressions to allow more diverse details. In this way, inter-individual differences and fine-grained deformations can be accurately modeled, which simultaneously takes into account the flexibility when applied to related tasks. Furthermore, inspired by linear blend skinning~\cite{lewis2000pose}, a Neural Blend-Field is presented to decompose the entire facial deformation or geometry into semantically meaningful regions encoded by a set of local implicit functions and adaptively blend them through a lightweight module, leading to more sophisticated representations with reduced parameters. Besides, a new preprocessing pipeline is designed, which bypasses the need of watertight face data as in existing SDF-based INR models and works well for various facial surfaces, \ie, either hardware-acquired or artificially synthesized.  
 
In summary, the main contributions of this study include:
\begin{itemize}
\vspace{-2mm}
\item We propose a novel INR-based 3DMM, which encodes complex face shape variations by two explicitly disentangled deformation fields, learning powerful representations in a fine-grained and semantically meaningful manner. 
\vspace{-2mm}
\item We present an effective preprocessing pipeline, which defines a general SDF for non-watertight 3D faces, enabling INRs to work with them for the first time. 
\vspace{-2mm}
\item We experimentally demonstrate that ImFace has the advantage in synthesizing high-quality 3D faces with plausible  details, outperforming the state-of-the-art counterparts in 3D face reconstruction. 
\end{itemize}

\begin{figure*}
  \centering
  \includegraphics[width=1\linewidth]{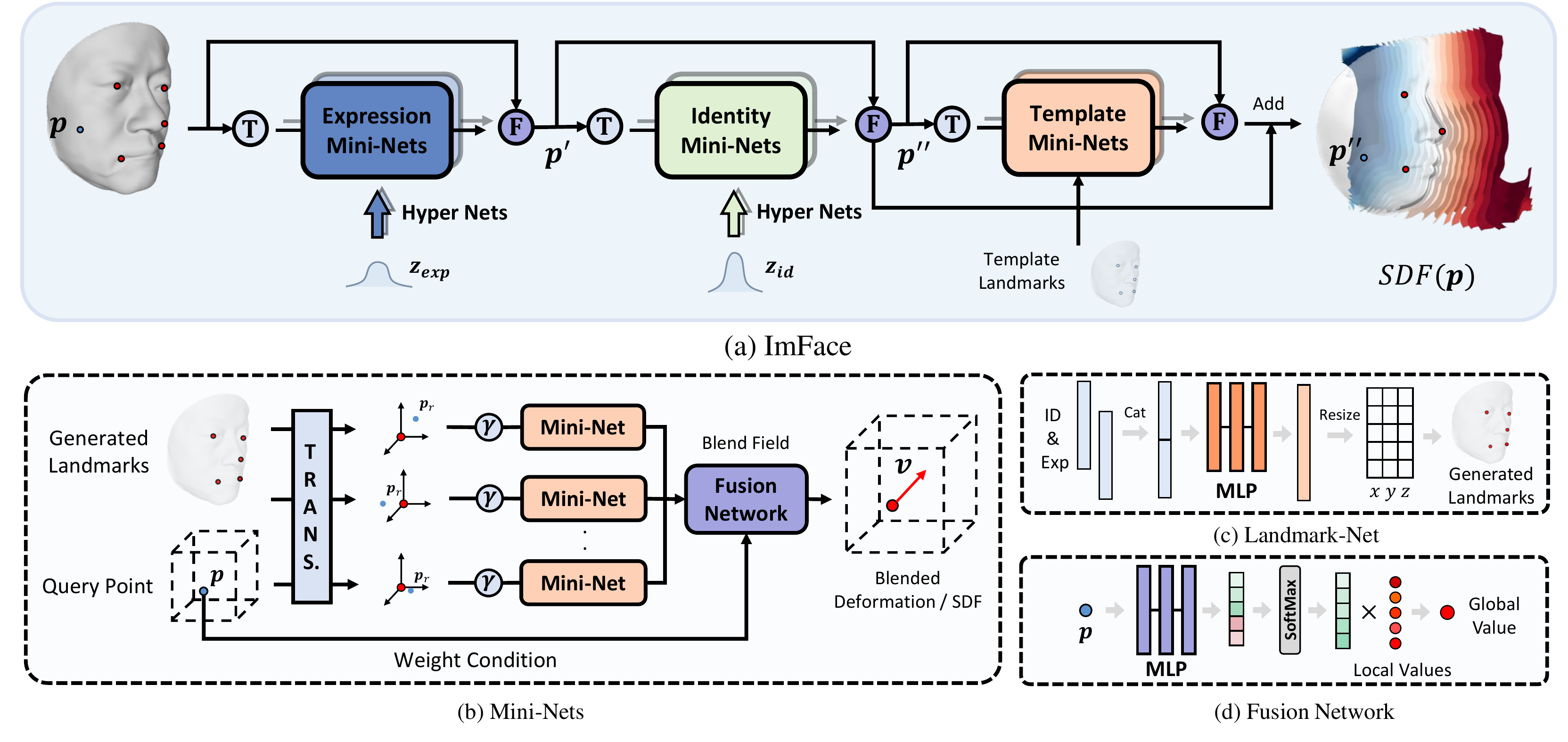}
  \vspace{-8mm}
  \caption{{\bf ImFace overview.} \textbf{(a)} The proposed network consists of three Mini-Nets blocks to explicitly disentangle shape morphs into separate deformation fields, where the Expression and Identity Mini-Nets blocks are associated with expression and identity deformations, respectively, and the Template Mini-Nets block learns the SDF of a template face space. \textbf{(b)} The Mini-Nets block is a shared architecture, which decomposes an entire facial feature into semantically meaningful parts and encodes them by a set of local field functions. It is tailed by a Fusion Network for more comprehensive representations. \textbf{(c)} The Landmark-Net is introduced to softly partition the entire facial surface. \textbf{(d)} The Fusion Network is a lightweight module conditioned on the query point position, which adaptively blends the local field functions, resulting in an elaborate Neural Blend-Field.}
  \label{fig:pipeline}
  \vspace{-3mm}
\end{figure*}

%-------------------------------------------------------------------------
\vspace{-2mm}
\section{Related Work}
\vspace{-1mm}

\para{3D Morphable Face Models} 3DMMs were first proposed by Blanz and
Vetter \cite{blanz1999morphable} as a general face representation. A known template mesh was registered to all training scans by utilizing Non-rigid Iterative Closest Point algorithm (NICP) \cite{amberg2007optimal}, and Principal Components Analysis (PCA) was used to span prior face distributions. To model identity-dependent expressions, 3DMMs were further extended to multilinear models\cite{vlasic2006face, brunton2014multilinear, bolkart2015groupwise}. After then, great advancements have been made with data improvement \cite{cao2013facewarehouse, booth2018large, li2017flame, yang2020facescape}. FLAME\cite{li2017flame} was an expressive model controlling facial expressions by combining jaw articulation with linear expression blendshapes. It was learned from a large 3D face dataset including D3DFACS\cite{cosker2011facs} and delivered more impressive results than ever, whereas nonlinear facial deformations cannot be well captured.

The development of deep networks has catalyzed more powerful nonlinear 3DMMs. A number of models were learned from 2D images \cite{tran2018nonlinear,tran2019towards,R_2021_CVPR}, but they mostly lacked high fidelity and fine details due to the low resolution of input images in this ill-posed inverse problem. To better leverage 3D face scans, Bagautdinov \textit{et al.} \cite{bagautdinov2018modeling} mapped 3D meshes to the 2D space and more studies \cite{cheng2019arxiv,ranjan2018generating,chen2021learning,bouritsas2019neural} directly learned 3DMMs from meshes by spectral or spiral convolutions. These neural networks were established upon discrete 3D representations, thus limiting the performance in the presence of complex deformations.

Please refer to \cite{egger20203d} for a more comprehensive report on 3DMMs. 

\para{Implicit Neural Representations} Recently, implicit neural functions have emerged as a more effective and suitable representation of 3D geometry~\cite{park2019deepsdf,mescheder2019occupancy,chen2019learning,icml2020_2086,pmlr-v139-lipman21a,sitzmann2019siren} as they model shapes continuously without discretization. To preserve fine details, the structured local features divided by shape elements\cite{genova2020local}, grids\cite{peng2020convolutional,ibing20213d} or octrees\cite{takikawa2021neural,tang2021octfield} were further exploited. Moreover, to well capture shape variations and correspondence relationships, an additional implicit deformation latent space was specially learned \cite{deng2021deformed,zheng2021deep}. However, high visual fidelity and variety could hardly be simultaneously achieved by existing techniques and they are therefore not so qualified to morphable face modeling. 
In addition, current studies on INRs mostly focused on the watertight input, such as the ones in ShapeNet~\cite{yi2016scalable}, and a number of methods on watertight human heads or bodies were proposed accordingly~\cite{corona2021smplicit,ramon2021h3d,yenamandra2021i3dmm,Alldieck_2021_ICCV,Chen_2021_ICCV,peng2021neural,saito2019pifu,saito2020pifuhd}. Among them, H3D-Net~\cite{ramon2021h3d} learned an implicit head shape space for 2D reconstruction, which was not a generic model. i3DMM~\cite{yenamandra2021i3dmm} was the first implicit 3D morphable model designed for human heads. However, it severely suffered from a low quality in representing facial regions. To overcome the limitation of watertightness, \cite{chibane2020ndf} learned an Unsigned Distance Function (UDF) to handle open surfaces, but the reconstructed results were not adequately plausible.

\section{Method}
We take advantages of INRs to learn a nonlinear 3D morphable face model. The proposed ImFace explicitly disentangles facial shape morphs into two separate deformation fields associated with identity and expression, respectively, and a deep SDF is learned to represent the template shape. All the fields are blended with a series of local implicit functions for more detailed representations.

\vspace{-1mm}
\subsection{Disentangled INRs Network}
\vspace{-1mm}

The fundamental idea of INRs is to train a neural network to fit a continuous function $f$, which implicitly represents surfaces through level-sets. The function can be defined in various formats, \eg occupancies~\cite{mescheder2019occupancy}, SDF~\cite{park2019deepsdf}, or UDF~\cite{chibane2020ndf}. We exploit a deep SDF conditioned on the latent embeddings of both expression and identity for comprehensive face representations. It outputs the signed distance $s$ from a query point:
\begin{equation}
  f:({\mathbf p}, {\mathbf z}_{exp}, {\mathbf z}_{id}) \in {\mathbb R}^{3} \times {\mathbb R}^{d_{exp}} \times {\mathbb R}^{d_{id}} \mapsto  s \in {\mathbb R},
\end{equation}
where ${\mathbf p} \in {\mathbb R}^3 $ is the coordinate of the query point in the 3D space, ${\mathbf z}_{exp}$ and ${\mathbf z}_{id}$ denote the 
expression and identity embeddings, respectively. 

Our goal is to learn a neural network to parameterize $f$, making it satisfy the genuine facial shape priors. As shown in Fig.~\ref{fig:pipeline}, the proposed network for Imface is composed of three Mini-Nets blocks, which explicitly disentangles the learning process of facial shape morphs, ensuring that inter-individual differences and fine-grained deformations can be accurately modeled. In particular, the first two Mini-Nets blocks learn separate deformation fields associated with expression- and identity-variation, respectively, and the Template Mini-Nets block learns a signed distance field of a template face shape. 

All the fields above are implemented by a shared Mini-Nets architecture, where the entire facial deformation or geometry is further decomposed into a number of semantically meaningful parts and encoded by a set of local field functions, so that rich details can be sufficiently captured. A lightweight module conditioned on the query point position, \textit{i.e.,} Fusion Network, is stacked at the end of the Mini-Nets block to adaptively blend the local fields. As such, an elaborate Neural Blend-Field is achieved. The three core components of ImFace work for different purposes and their structures are slightly changed accordingly. We briefly describe them as follows: 

\para{Expression Mini-Nets (ExpNet)} The facial deformations incurred by expressions are represented by ExpNet $\mathcal{E}$, which learns an observation-to-canonical warping for every face scan:
\begin{equation}
  \mathcal{E}:({\mathbf p}, {\mathbf z}_{exp}, l)  \mapsto  {\mathbf p}' \in {\mathbb R}^{3},
\end{equation}
\noindent where $l \in {\mathbb R}^{k \times 3}$ denotes $k$ 3D landmarks on an observed face generated by a Landmark-Net $\eta: ({\mathbf z}_{exp}, {\mathbf z}_{id}) \mapsto l $, introduced to localize the query point ${\mathbf p}$ in the Neural Blend-Field. The point ${\mathbf p}$ in the observation space is deformed by $\mathcal{E}$ to a new point ${\mathbf p}'$ in the person-specific canonical space, which represents faces with a neutral expression.

\para{Identity Mini-Nets (IDNet)} To model shape morphs among individuals, IDNet $\mathcal{I}$ further warps the canonical space to a template shape space shared by all faces:
\begin{equation}
  \mathcal{I}:({\mathbf p}', {\mathbf z}_{id}, l')  \mapsto  ({\mathbf p}'', \delta) \in {\mathbb R}^{3} \times {\mathbb R},
\end{equation}
\noindent where $l' \in {\mathbb R}^{k \times 3}$ denotes $k$ landmarks on the canonical face generated by another Landmark-Net conditioned only on the identity embedding $\eta': {\mathbf z}_{id} \mapsto l' $, and ${\mathbf p}''$ is the deformed point in the template space. To cope with the possible non-existent correspondences generated during preprocessing, $\mathcal{I}$ additionally predicts a residual term $\delta \in {\mathbb R}$ to correct the predicted SDF value $s_0$, which is similar to \cite{deng2021deformed}.

\para{Template Mini-Nets (TempNet)} TempNet $\mathcal{T}$ learns a signed distance field of the shared template face:
\begin{equation}
  \mathcal{T}:({\mathbf p}'', l'')  \mapsto  s_0 \in {\mathbb R},
\end{equation}
where \noindent $l'' \in {\mathbb R}^{k \times 3}$ denotes $k$ landmarks on the template face, which is averaged on the whole training set, and $s_0$ is the uncorrected SDF value.  The final SDF value of a query point is calculated via $s = s_0 + \delta$, and the ImFace model can be ultimately formulated as: %\vspace{-0.3cm}
{\small
\begin{equation}
  f({\mathbf p}) = \mathcal{T}(\mathcal{I}_{{\mathbf p}''}(\mathcal{E}({\mathbf p}, {\mathbf z}_{exp}),{\mathbf z}_{id})) + \mathcal{I}_{\delta}(\mathcal{E}({\mathbf p}, {\mathbf z}_{exp}),{\mathbf z}_{id}).
\end{equation}
}
\hspace{-0.5em} The proposed ImFace learns face morphs by disentangled deformation fields in a fine-grained and meaningful manner, ensuring that more diverse and sophisticated facial deformations can be accurately learned. We detailedly introduce the architectures of the main modules, learning strategy, training critics and data preprocessing pipeline in the subsequent.

\vspace{-1mm}
\subsection{Neural Blend-Field}
\label{subsec:nbf}
\vspace{-1mm}
The Mini-Nets block is a common architecture shared by the three sub-networks $\mathcal{E}$, $\mathcal{I}$, and $\mathcal{T}$. It learns a continuous field function ${\boldsymbol \psi}:{\mathbf x} \in {\mathbb R}^3 \mapsto v $, to produce a Neural Blend-Field for comprehensive face representations. In particular, to overcome the limited expressivity of a single network, we decompose a face space into a set of semantically meaningful local regions, and learn $v$ (\eg deformation or signed distance value) individually before blending. Such design is inspired by the recent INRs study~\cite{peng2021animatable} on human body, which introduces the linear blend skinning algorithm~\cite{lewis2000pose} to make the network learn from separate transformations of body parts. To better represent detailed facial surfaces, we replace the constant transformation term in the original linear blend skinning algorithm with ${\boldsymbol \psi}_n({\mathbf x}-l_n)$, and define the Neural Blend-Field as:
\vspace{-2mm}
\begin{equation}
  v={\boldsymbol \psi}({\mathbf x}) =\sum_{n=1}^k{{\it w}_n({\mathbf x}) {\boldsymbol \psi}_n({\mathbf x}-l_n)},
\vspace{-2mm}
\end{equation}
\noindent where $l_n$ is a parameter that describes the $n$-th local region, ${\it w}_n({\mathbf x})$ is the $n$-th blend weight, and ${\boldsymbol \psi}_n({\mathbf x}-l_n)$ is the corresponding local field. In this way, the blending is performed on a series of local fields, rather than calculating a weighted average of the output values $v$ of some fixed positions, leading to more powerful representation capability in handling complicated local features.

Specifically, five landmarks located at the outer eye corners, mouth corners, and nose tip are utilized to describe the local regions $(l_n \in {\mathbb R}^{3})_{n=1}^5$, and each region is assigned a tiny MLP with sinusoidal activations~\cite{sitzmann2019siren} to generate the local field, denoted as ${\boldsymbol \psi}_n$. To capture high-frequency local variations, we leverage sinusoidal positional encoding $\gamma$~\cite{mildenhall2020nerf} on the coordinate ${\mathbf x}-l_n$. At the end of a Mini-Nets block, a lightweight Fusion Network conditioned on the absolute coordinate of input ${\mathbf x}$ is equipped, which is implemented by a 3-layer MLP with softmax to predict the blend weights $(w_n \in {\mathbb R}^+)_{n=1}^5$.

\para{Deformation Formulation} We formulate the deformation with a SE(3) field $({\boldsymbol \omega},{\mathbf v}) \in {\mathbb R}^{6} $, where ${\boldsymbol \omega} \in so(3)$ is a rotate vector representing the screw axis and the angle of rotation. The deformed coordinates ${\mathbf x}'$ can be calculated by ${e^{\boldsymbol \omega}}{\mathbf x}+{\mathbf t}$, where the rotation matrix ${e^{\boldsymbol \omega}}$ (exponential map form of Rodrigues’ formula) is written as:
{\small
\begin{equation}
  {e^{\boldsymbol \omega}}={\bf I}+ \frac{\sin \left \| {\boldsymbol \omega} \right \|}{\left \| {\boldsymbol \omega} \right \|} {\boldsymbol \omega}^{\wedge} + \frac{1-\cos \left \| {\boldsymbol \omega} \right \|}{{\left \| {\boldsymbol \omega} \right \|}^2} ({\boldsymbol \omega}^{\wedge})^2,
\end{equation}}
\hspace{-0.5em} and the translation ${\mathbf t}$ is formulated as:
{\small
  \begin{equation}
    {\mathbf t}=\left[{\bf I}+ \frac{1-\cos \left \| {\boldsymbol \omega} \right \|}{{\left \| {\boldsymbol \omega} \right \|}^2} {\boldsymbol \omega}^{\wedge} + \frac{\left \| {\boldsymbol \omega} \right \|-\sin \left \| {\boldsymbol \omega} \right \|}{{\left \| {\boldsymbol \omega} \right \|}^3} ({\boldsymbol \omega}^{\wedge})^2 \right]{\mathbf v},
  \end{equation}
}
\hspace{-0.39em}where ${\boldsymbol \omega}^{\wedge}$ denotes the skew-symmetric matrix of ${\boldsymbol \omega}$. We exploit SE(3) to describe facial shape morphs for its superior capability in handling mandibular rotations and higher robustness to pose perturbations than the common translation deformation
${\mathbf x}'={\mathbf x}+{\mathbf t}$.

\para{Hyper Nets} To obtain a more compact and expressive latent space, we then introduce a meta-learning approach\cite{sitzmann2019siren}. A Hyper Net $\phi_n$ is implemented by an MLP and it predicts the instance-specific parameters for ExpNet $\mathcal{E}$ and IDNet $\mathcal{I}$. It takes a latent code ${\mathbf z}$ as input and generates the parameters for the neurons in a Mini-Net ${\boldsymbol \psi}_n$ so that the learned facial representations possess a higher variety.

\begin{figure*}
  \centering
  \begin{subfigure}{0.33\linewidth}
    \includegraphics[width=1\linewidth]{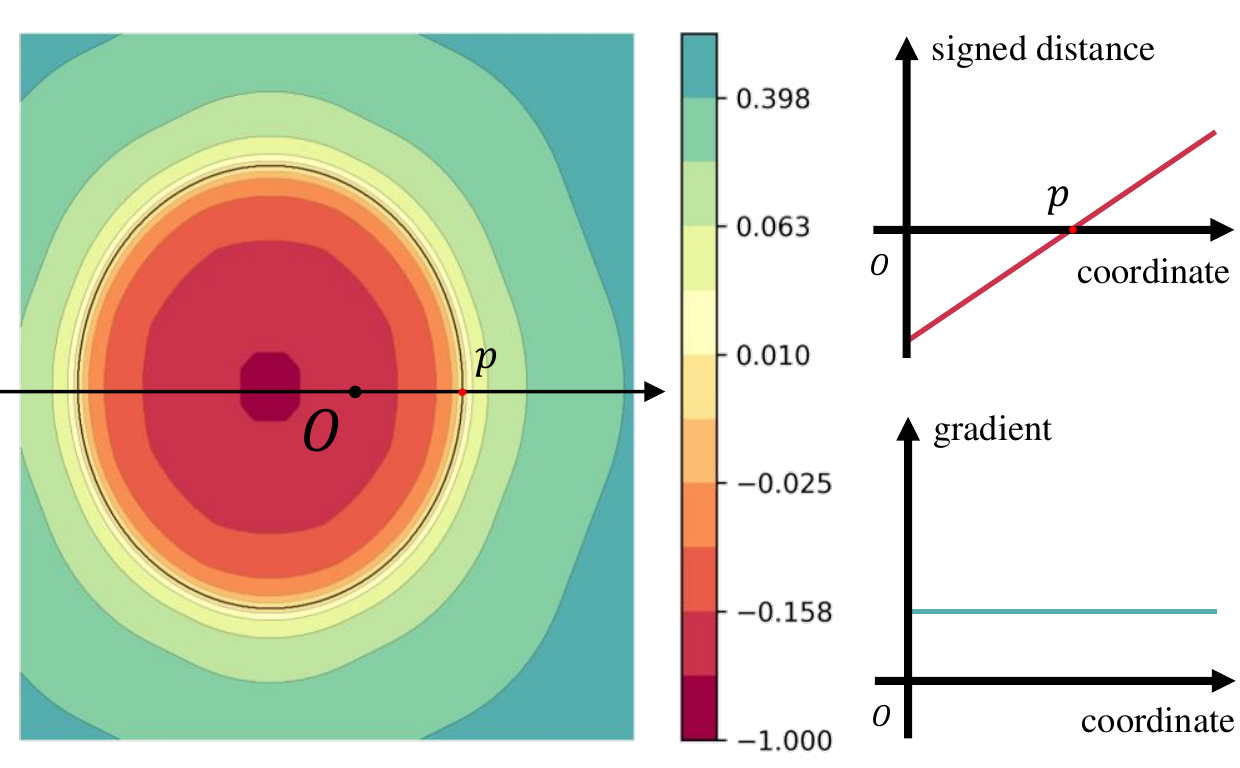}
    \caption{Signed Distance Field}
  \end{subfigure}
  \hfill
  \begin{subfigure}{0.33\linewidth}
    \includegraphics[width=1\linewidth]{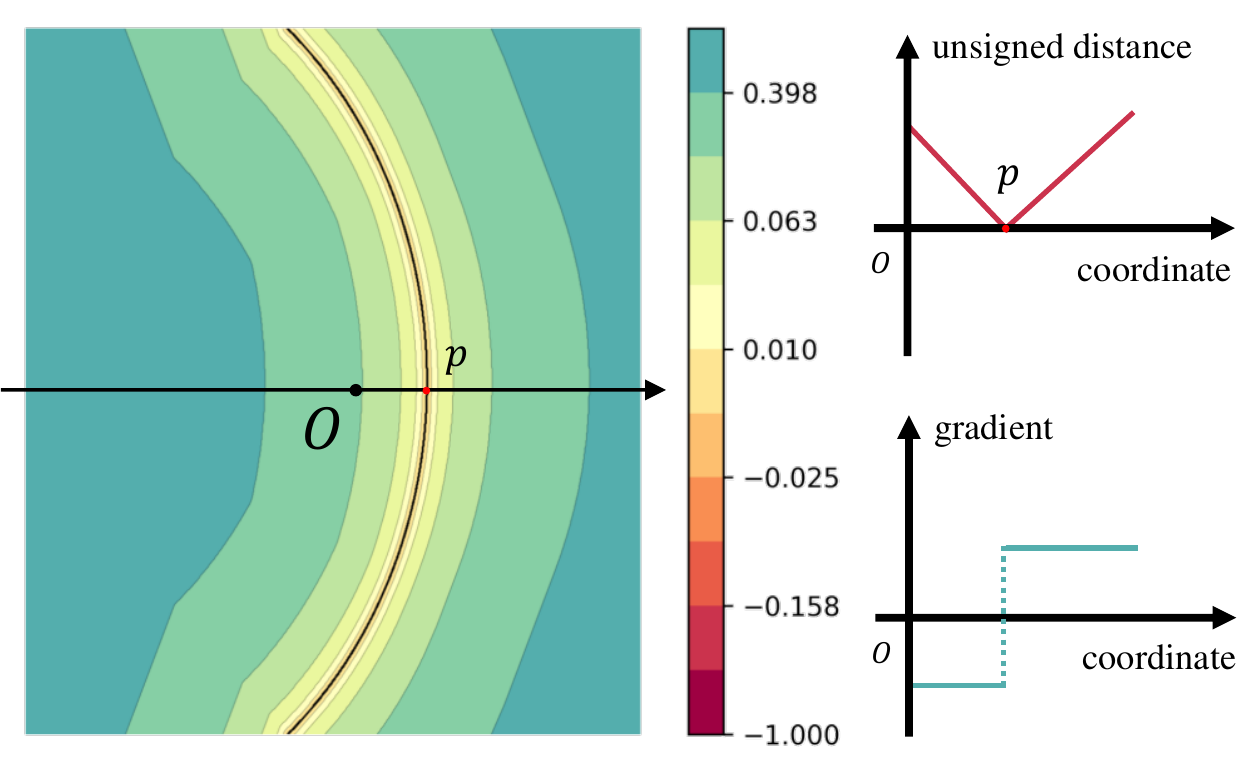}
    \caption{Unsigned Distance Field}
  \end{subfigure}
  \hfill
  \begin{subfigure}{0.33\linewidth}
    \includegraphics[width=1\linewidth]{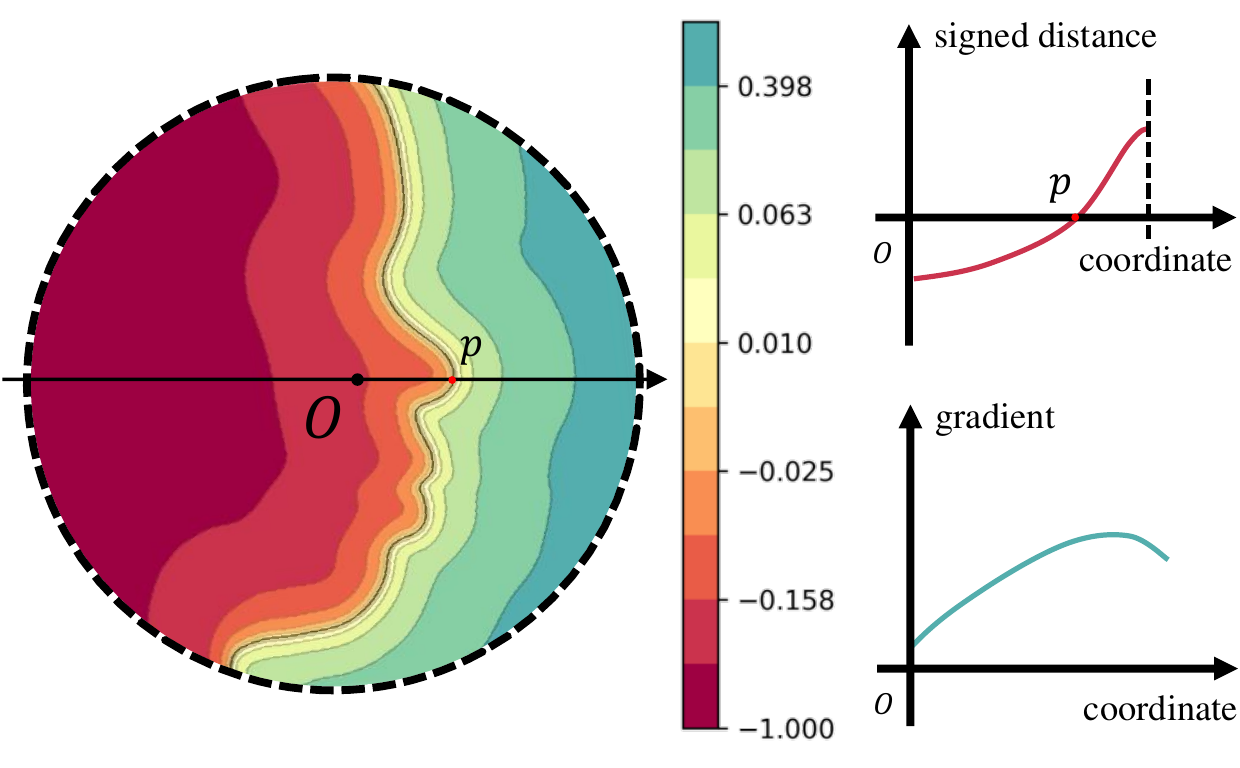}
    \caption{SDF on pseudo watertight face}
  \end{subfigure}

  \vspace{-3mm}
  \caption{\textbf{(a)} SDF is able to represent closed shapes. \textbf{(b)} UDF is capable of representing an open surface but the gradient is discontinuous at the boundary, making it hard to be fitted by neural networks. \textbf{(c)} The proposed method generates pseudo watertight faces and restricts implicit functions on them, enabling implicit neural networks to learn geometry representations on 3D faces.}
  \label{fig:wt}
  \vspace{-3mm}
\end{figure*}

\vspace{-1mm}
\subsection{Improved Expression Embedding Learning}
\label{subsec:emb}
\vspace{-1mm}
The auto-decoder framework proposed by \cite{park2019deepsdf} has been widely adopted in INRs to jointly learn embeddings and network parameters. In the previous attempt \cite{yenamandra2021i3dmm}, each expression type is encoded by one embedding for attribute disentangling. Unfortunately, such an embedding is merely able to represent the average shape morph of an expression type, making the learned latent space fail to capture more diverse deformation details across individuals. To avoid the dilemma above, we improve the learning strategy by treating each non-neutral face scan as a unique expression and generating a specific embedding for it. In this way, the latent space is significantly extended, which enables $\mathcal{E}$ to represent more fine-grained details. On the other side, there exists a potential failure mode that the identity properties are tangled into the expression space again, and $\mathcal{I}$ collapses to an identity mapping. To tackle this challenge, we suppress $\mathcal{E}$ when the current training sample is a neutral face, written as:
\begin{equation}
  \mathcal{E}({\mathbf p_{nu}}, {\mathbf z}_{exp}, l)\equiv{\mathbf p}_{nu},
\end{equation}
where ${\mathbf p_{nu}}$ denotes a point from a neutral face. By applying such a learning strategy, $\mathcal{I}$ and $\mathcal{T}$ jointly learn shape representations on neutral faces, and $\mathcal{E}$ focuses only on expression deformations. Moreover, only neutral labels are required during training, bypassing the dense expression labels.
\vspace{-1mm}
\subsection{Loss Functions}
\vspace{-1mm}
ImFace is trained with several loss functions to learn plausible facial shape representations and dense correspondence.

\para{Reconstruction Loss} The basic SDF structure loss is applied to learn implicit fields:
{\small
  \begin{equation}
    {\cal L}_{sdf}^i= 
    {\lambda_1} \sum_{{\mathbf p} \in \Omega_i} \left| f({\mathbf p})-{\bar s} \right|
    + {\lambda_2} \sum_{{\mathbf p} \in \Omega_i} (1-\left \langle {\nabla f({\mathbf p})} , {\mathbf{\bar n}} \right \rangle),
  \end{equation}
}
\hspace{-0.38em}where ${\bar s}$ and ${\mathbf{\bar n}}$ denote the ground-truth SDF values and the field gradients, respectively. ${\Omega_i}$ is the sampling space of the face scan $i$, and $\lambda$ indicates the trade-off parameter.

\para{Eikonal Loss} To obtain reasonable fields throughout the network, multiple Eikonal losses are used to enforce the L-2 norm of spatial gradients to be unit:
{\small
  \begin{equation}
    \begin{aligned} 
      {\cal L}_{eik}^i \! = &
      {\lambda_3} \! \sum_{{\mathbf p} \in \Omega_i} \! \big(
        | \| \nabla f({\mathbf p})\| \!-\!1 |
        \!+\! | \| \nabla \mathcal{T}(\mathcal{I}({\mathbf p'}))\| \!-\!1 | 
      \big),
    \end{aligned}
  \end{equation}
}
\hspace{-0.38em}where ${\cal L}_{eik}^i$ enables the network to satisfy the Eikonal constraint\cite{icml2020_2086} in the observation and canonical spaces simultaneously, which also contributes to a reasonable correspondence along face deformations at all the network stages.

\para{Embedding Loss} It regularizes the embedding with a zero-mean Gaussian prior: 
{\small
\begin{equation}
    {\cal L}_{emb}^i = 
    {\lambda_4} \left( {\|{\mathbf z}_{exp}\|}^2+{\|{\mathbf z}_{id}\|}^2 \right).
\end{equation}}

\para{Landmark Generation Loss} The ${\mathit l}_1$-loss is used to learn the Landmark-Nets $\eta$, $\eta'$:
{\small
\begin{equation}
  {\cal L}_{lmk_g}^i\!=\!{\lambda_5} 
  \sum_{n=1}^k \left( |l_n-{\bar {l_n^i}}|+|l'_n-{\bar {l_n'}}| \right),
\end{equation}
}
\hspace{-0.38em}where ${\bar {l^i}}$ denotes the $k$ labeled landmarks on sample $i$, and ${\bar {l'}}$ denotes the landmarks on the corresponding neutral face.

\para{Landmark Consistency Loss} We exploit this loss to guide the deformed landmarks to be located at the corresponding positions on the ground-truth neutral and template faces for better correspondence performance: 
{\small
\begin{equation}
  {\cal L}_{lmk_c}^i\!=\!{\lambda_6} 
  \sum_{n=1}^{k} \left( |\mathcal{E}(l_n)-{\bar {l'}}_n|+|\mathcal{I}(\mathcal{E}(l_n))- {l_n''}| \right).
\end{equation}
}

\begin{figure*}
  \centering
  \includegraphics[width=1\linewidth]{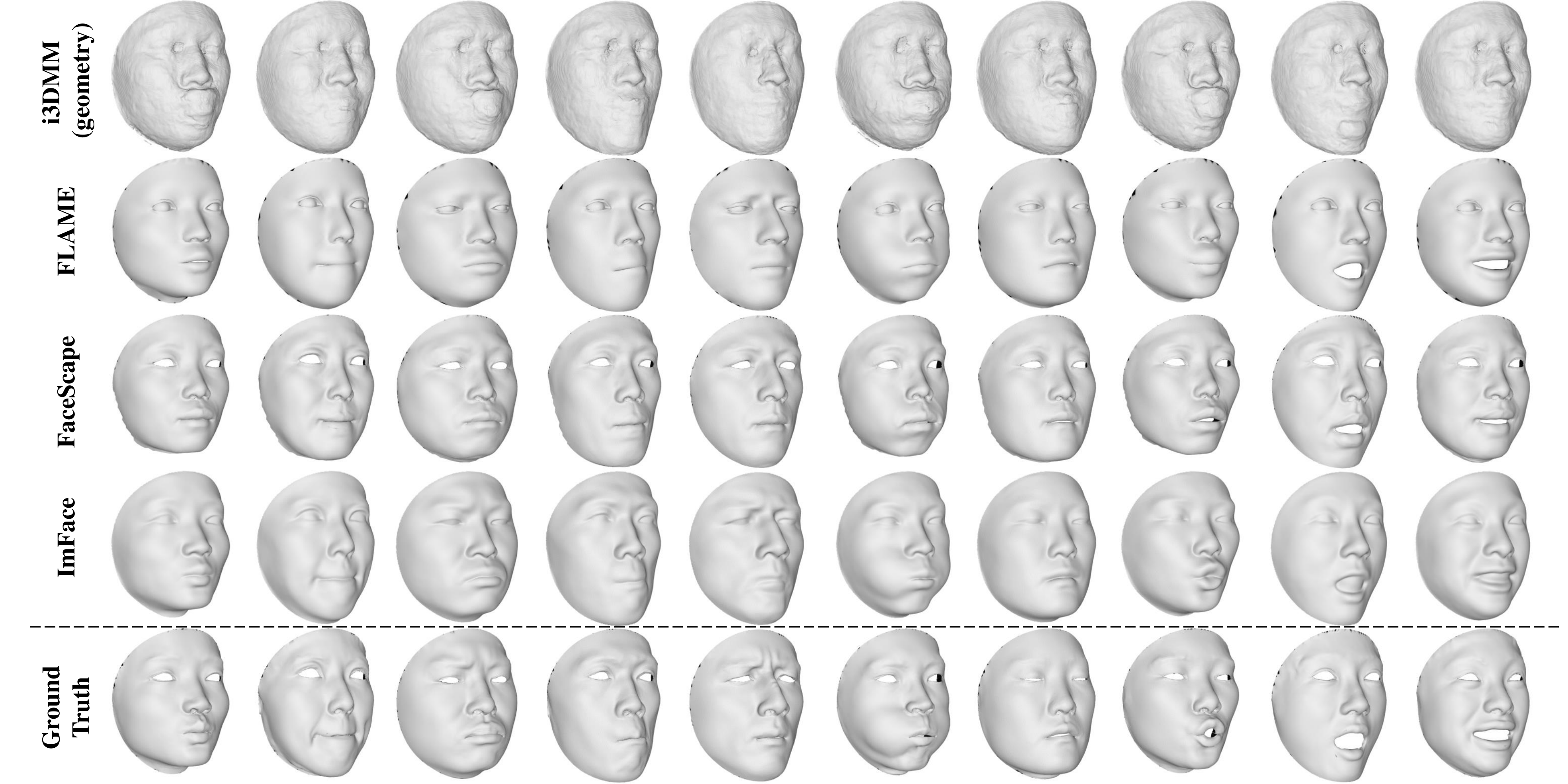}
   \vspace{-7mm}
  \caption{Reconstruction comparison with i3DMM\cite{yenamandra2021i3dmm}, FLAME\cite{li2017flame}, and FaceScape\cite{yang2020facescape}. Each column corresponds to a person with a non-neutral expression. Visually inspected, ImFace captures much richer shape variations with more compact latent embeddings.}
  \label{fig:experiment_recon}
  \vspace{-2mm}
\end{figure*}

\para{Residual Constraint}
As in \cite{deng2021deformed}, to avoid that the residual item $\delta$ learns too much template face information and downgrades the morphable model, we penalize $\delta$ by:
{\small
  \begin{equation}
    {\cal L}_{res}^i= 
    {\lambda_7} \sum_{{\mathbf p} \in \Omega_i} |\delta({\mathbf p})|.
  \end{equation}
}
\hspace{-0.5em} The total training loss is calculated on all face samples indexed by $i$,  finally formulated as:
{\small
  \begin{equation}
    {\cal L} = \sum_i ({\cal L}_{sdf}^i+{\cal L}_{eik}^i+{\cal L}_{emb}^i+{\cal L}_{lmk_g}^i+{\cal L}_{lmk_c}^i+{\cal L}_{res}^i).
  \end{equation}
}
\hspace{-0.5em} At the testing phase, for each 3D face indexed by $j$, we minimize the following objective to obtain its latent embedding and the reconstructed 3D face:
{\small
  \begin{equation}
    \mathop{\arg\min}\limits_{{\mathbf z}_{exp},{\mathbf z}_{id}} \sum_j ({\cal L}_{sdf}^j+{\cal L}_{eik}^j+{\cal L}_{emb}^j).
  \label{fit}
  \end{equation}
}

\vspace{-2mm}
\subsection{Data Preprocessing}
\vspace{-1mm}
\label{subsec:process}

Since neural networks excel in fitting functions that are differentiable everywhere, the current studies on implicit functions generally require watertight input. Although functions like UDF do not demand watertightness, they are nondifferentiable when crossing a surface and are not so competent at handling details (see Fig.~\ref{fig:wt} for an illustration). We present an effective preprocessing pipeline, which generates pseudo watertight faces and defines a general SDF on them, so that the geometry and correspondence can be learned as exquisitely as on watertight objects. 

\para{Pseudo watertight face generation} The faces are rigidly aligned to frontal using landmarks and each mesh is normalized to a unit of 10 $cm$. The coordinate origin is set at the point 4 $cm$ behind the nose tip, and the sphere with a radius of 10 $cm$ is then defined as the sampling area where the mesh triangles outside are cropped away. The Ray-Triangle Intersection Algorithm\cite{moller1997fast} is applied to remove the hidden surface such as nasal and oral cavity, and the Delaunay Triangulation Algorithm~\cite{lee1980two} is then performed on x-y coordinates for an oriented and pseudo watertight mesh. 

\para{SDF computation on facial surfaces} 
With the pseudo watertight faces generated, SDF values can then be computed through a distance transform on them. The sign of the samples is determined simply by the angle between its distance vector to the nearest surface and the z-axis positive direction. The values of coordinates behind the facial surface are defined to be negative. We uniformly sample 250,000 points on each facial surface and 15,000 points in the sphere and calculate their signed distance and gradient vectors. The sampled data are eventually formulated as $\{({\bf p},{\bf \bar{n}},\bar{s})\}$ triplets (query point, gradient vectors, signed distance value) for ImFace training.

\section{Experiments}
We conduct extensive subjective and objective evaluations on ImFace, and ablation studies are performed to validate the specifically
designed modules.

\begin{figure*}
  \centering
   \includegraphics[width=1\linewidth]{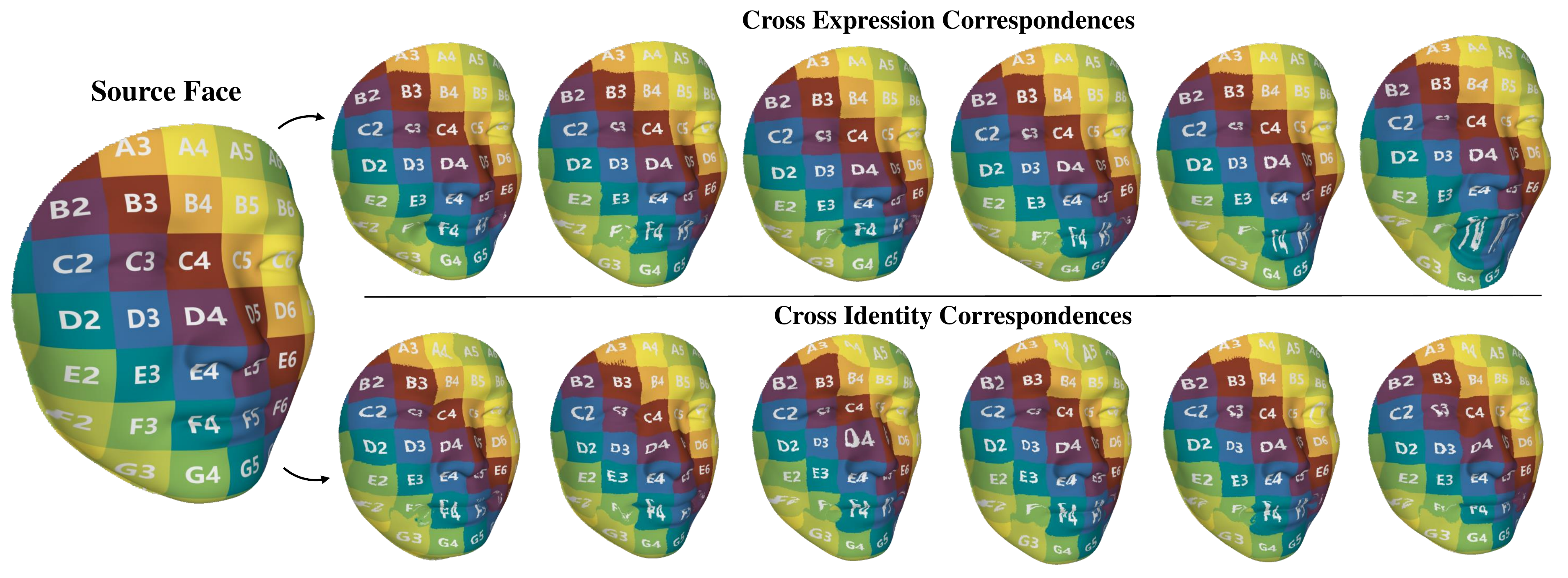}
    \vspace{-8mm}
   \caption{Correspondence results. The leftmost face is morphed into multiple expressions (upper row) and identities (bottom row).}
   \label{fig:corr}
   \vspace{-2mm}
\end{figure*}

\para{Dataset} FaceScape\cite{yang2020facescape} is a large-scale high quality 3D face Dataset consisting of 938 individuals with 20 types of expressions. The data from 365 individuals are publicly available and we mainly use them for experiments. To be specific, 5,323 face scans from 355 persons with 15 expressions are sampled as the training set, and another 200 face scans from the remaining 10 persons with 20 expressions are used as the testing set. 

\para{Network Architecture} All Mini-Nets blocks ${\boldsymbol \psi}_n$ are implemented as MLPs with 3 hidden layers and 32 dimensional hidden features activated by sine. The Hyper Nets $\phi_n$ are 3-layer MLPs activated by ReLU, where the hidden layer dimensionality is 64. Landmark-Nets $\eta$ and $\eta'$ have three 128 dimensional fully connected layers. Please refer to the supplementary material for more details of the networks.

\para{Implementation Details} The model is trained with Adam in an end-to-end manner. We train the model for 1,500 epochs with an initial learning rate of 0.0001, and after 200 epochs, we decay it by a factor of 0.95 for every 10 epochs. The training phase takes around 2 days on 4 NVIDIA RTX 3090 GPUs with minibatches of size 72. During testing, it takes about 4 hours to optimize 200 samples on a single GPU.

\subsection{Reconstruction}
We use the proposed ImFace model to fit face scans by optimizing Eq.~(\ref{fit}) and compare the reconstruction results with FLAME\cite{li2017flame}, FaceScape\cite{yang2020facescape} and the geometry model of i3DMM\cite{yenamandra2021i3dmm}, which signify the state-of-the-art. The official code of FLAME is used to fit the full face scans in the test set, with 300 identity parameters and 100 expression parameters. For FaceScape, we use their released bilinear model built from 938 individuals for testing, where the identity and expression parameters are 300 and 52, respectively. Note that our test scans are included in the training set of FaceScape. In addition, we modify its official code to fit the full scans instead of only the landmarks for improved results. For i3DMM, since the original model is trained only on 58 individuals, we therefore re-train the model on the same training set as in ImFace for fair comparison. In both i3DMM and ImFace, the identity and expression embeddings are 128-dimensional. 

\begin{table}
\small
  \centering
  \begin{tabular}{@{}cccc@{}}
    \toprule
    Metrics & \makecell[c]{Dim.} & \makecell[c]{Chamfer($mm$)$^{\dag}$} &  \makecell[c]{F-score@0.001$^{\P}$} \\
    \midrule
    i3DMM\cite{yenamandra2021i3dmm} & 256 & 1.635 & 42.26\\
    FLAME\cite{li2017flame} & 400 & 0.971 & 64.73\\
    FaceScape\cite{yang2020facescape} & 352 & 0.929 & 67.09\\
    ImFace & 256 & \bf{0.625} & \bf{91.11}\\
    \bottomrule
  \end{tabular}
  \caption{Quantitative comparison with the state-of-the-art methods ($^{\dag}$Lower is better; $^{\P}$Higher is better).}
  \label{table:quantitative}
  \vspace{-0.3cm}
\end{table}

\para{Qualitative Evaluation} Fig. \ref{fig:experiment_recon} visualizes the reconstruction results achieved by different models, where each column corresponds to a test person with a non-neutral expression. The results also include unseen expressions during learning. i3DMM is the first deep implicit model for human heads, but it is less capable of capturing complicated deformations and fine-grained details under a relatively intricate circumstance, resulting in artifacts on the reconstructed faces. FLAME is able to well present identity characteristics, but is not so competent at dealing with nonlinear deformations, delivering stiff facial expressions. FaceScape performs more favourably mainly due to the high-quality training scans and the test faces are included in the training set, but it still cannot precisely present expression morphs. Comparatively, ImFace reconstructs faces with more accurate identity and expression properties, and it is able to preserve subtle and rich nonlinear facial muscle deformations such as frowns and pouts by fewer latent parameters.

\para{Quantitative Evaluation} To make fair comparison, all faces are processed as described in Sec.~\ref{subsec:process} to remove the inner structures like eyeballs, so that the quantitative metrics can be computed in the same facial region for all the models. Specifically, the symmetric Chamfer distance and F-score are used as metrics, and the threshold of F-score is set to 0.001 as a strict standard. The results are shown in Table~\ref{table:quantitative}. As we can see, ImFace exceeds the counterparts by a large margin under both the metrics, which clearly validates its effectiveness.

\begin{figure}
  \centering
  \setlength{\abovecaptionskip}{0pt}
  \setlength{\belowcaptionskip}{0pt}
  \includegraphics[width=0.9\linewidth]{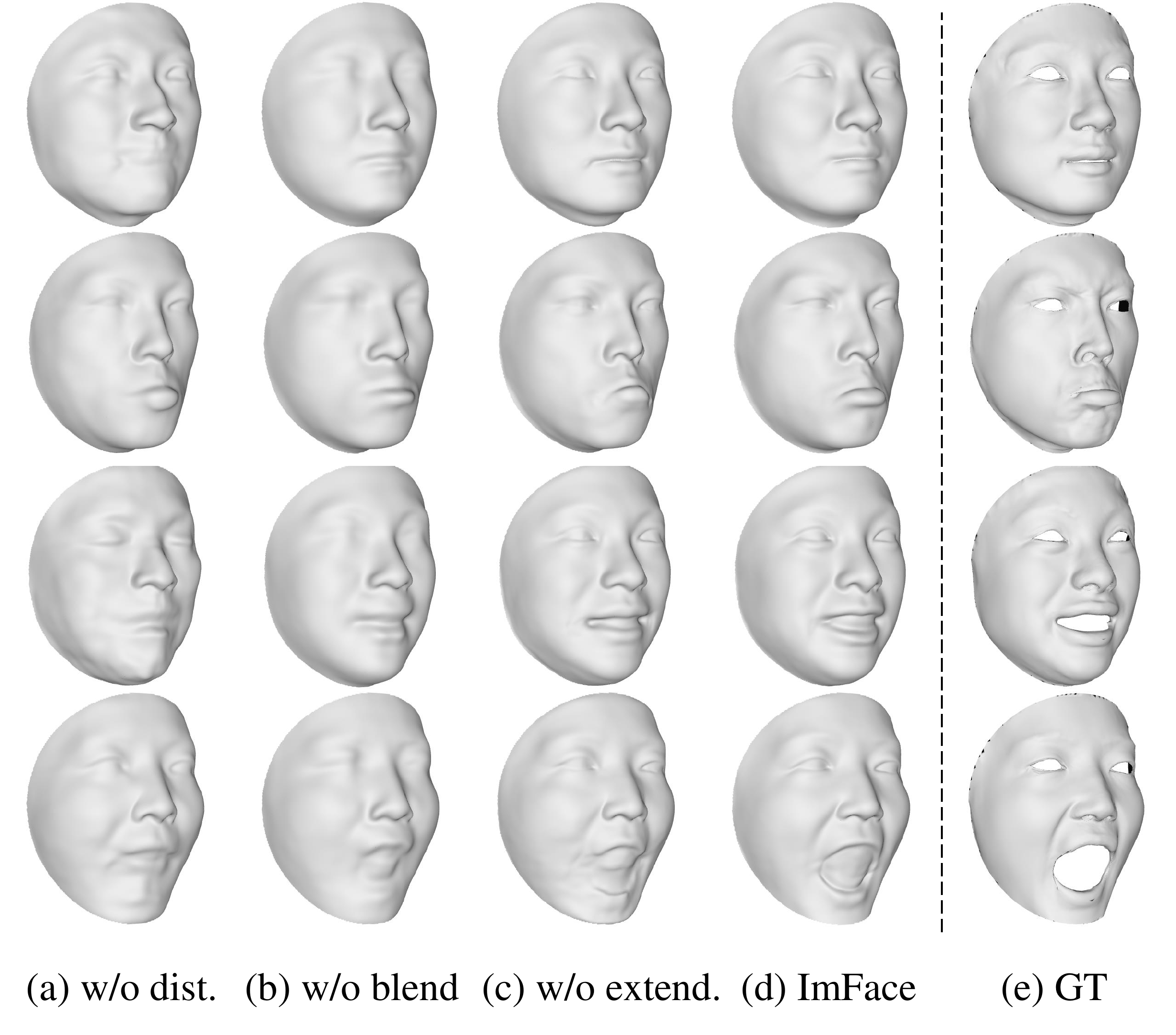}
   \caption{Qualitative ablation study results.}
   \label{fig:ablation}
\vspace{-2mm}
\end{figure}

\subsection{Correspondence}
In contrast to existing methods that generally require accurate face registration, correspondences can be automatically learned in INRs models. We further design training critic to enhance such a feature, and this evaluation aims to confirm it. Given two 3D faces, we use ImFace to fit them and deform the dense sampled points to the template space so that point-to-point correspondences can be fulfilled by nearest neighbor search. Fig.~\ref{fig:corr} visualizes some results generated by our method, where we manually paint color patterns on the shapes to better check the quality. In general, ImFace is able to establish pleasing correspondences across various expressions and identities. Meanwhile, it can be inspected that tiny internal texture dispersion indeed occasionally occurs around mouth corners, and this is mainly because facial shapes change drastically in these local areas under different expressions, which is also very hard case to specialized correspondence methods.

\begin{table}
  \centering
  \setlength{\belowcaptionskip}{0pt}
  \begin{tabular}{@{}ccc@{}}
    \toprule
    Metrics & Chamfer ($mm$) $^{\dag}$ &  F-score@0.001 $^{\P}$ \\
    \midrule
    Ours w/o dist. & 0.772 & 82.70\\
    Ours w/o blend & 0.767 & 82.37\\
    Ours w/o extend. & 0.705 & 86.98\\
    ImFace & \bf{0.625} & \bf{91.11}\\
    \bottomrule
  \end{tabular}
   \vspace{-2mm}
  \caption{Quantitative ablation study results. }
  \label{table:ablation}
  \vspace{-4mm}
\end{table}

\subsection{Ablation Study}
ImFace is built on the following core components: disentangled deformation fields (dist.), Neural Blend-Field (blend), and improved expression embedding learning (extend.). We experimentally verify the credits of such designs. 

\para{On Disentangled Deformation Fields} To highlight the disentangled deformation learning process, we build a baseline network which contains only one deformation field to learn face shape morphs universally. Accordingly, ${\mathbf z}_{exp}$ and ${\mathbf z}_{id}$ are concatenated as the input of the hyper net. Fig.~\ref{fig:ablation} (a) provides a demonstration. In spite of some fine-grained details brought by other designs, there exists a chaos on the reconstructed faces, especially for the ones with large expressions. The quantitative results in Table~\ref{table:ablation} also indicate the significance of decoupled deformation learning.

\para{On Neural Blend-Field} We replace the Neural Blend-Field in $\mathcal{E}$, $\mathcal{I}$, $\mathcal{T}$ with vanilla MLPs of the same amount of parameters, which directly predict global deformations or SDF values of an entire face. As shown in Fig.~\ref{fig:ablation} (b), a visible blur appears due to the limited capability in learning high-frequency details. The quantitative results in Table~\ref{table:ablation} confirm the necessity of Neural Blend-Field in learning sophisticated representations.

\para{On Improved Embedding Learning} This strategy is introduced to learn more diverse and fine-grained facial deformations. As shown in Fig.~\ref{fig:ablation} (c), when restricting the number of expression embeddings to be the same as expression categories, the generated expressions tend to be average. Moreover, for exaggerated expressions, such as mouth stretch, the counterpart models can hardly converge to a reasonable state.

\vspace{-1mm}
\section{Discussion}
\para{Limitations} Despite the significant advance in 3D facial shape representation, we mainly focus on face geometry modeling, whereas facial texture is less considered. Indeed, a basic texture model can be achieved by plugging a color field, but a more comprehensive INR-based 3DMM which presents the facial appearance with realistic diffuse and specular albedo remains to be explored.

\para{Societal Impact} Our model aims at high-quality face modeling, and similar to existing models, it has the potential to be applied to downstream scenarios such as 2D reconstruction and face animation, which may result in unethical practices like privacy invasion or identity fraud. We encourage researchers and developers to consider the questions, such as how to prevent personal face data from being maliciously accessed, before applying the model to real world.

\vspace{-1mm}
\section{Conclusion}
This paper presents a novel nonlinear 3D morphable face model with INRs, \ie ImFace, which learns complex facial  shape variations by two explicitly disentangled deformation fields associated with expression and identity, respectively, along with an improved embedding learning strategy to allow more fine-grained expressions. To precisely capture detailed facial deformations and geometries, it further presents a Neural Blend-Field. In addition, an effective preprocessing pipeline is presented, which enables INRs to work with non-watertight facial surfaces for the first time. Experiments show that ImFace is competent for this issue and outperforms the state-of-the-art counterparts.

\vspace{-1mm}
\section*{Acknowledgment}
This work is partly supported by the National Natural Science Foundation of China (No. 62022011), Beijing Municipal Natural Science Foundation (No. 4222049), the Research Program of State Key Laboratory of Software Development Environment (SKLSDE-2021ZX-04), and the Fundamental Research Funds for the Central Universities.

%%%%%%%%% REFERENCES
{\small
\bibliographystyle{ieee_fullname}
\bibliography{egbib}

\begin{thebibliography}{10}\itemsep=-1pt

\bibitem{pyrender}
\url{https://github.com/mmatl/pyrender}.

\bibitem{abrevaya2018spatiotemporal}
Victoria~Fern{\'a}ndez Abrevaya, Stefanie Wuhrer, and Edmond Boyer.
\newblock Spatiotemporal modeling for efficient registration of dynamic 3d
  faces.
\newblock In {\em 3DV}, 2018.

\bibitem{aldrian2012inverse}
Oswald Aldrian and William~AP Smith.
\newblock Inverse rendering of faces with a 3d morphable model.
\newblock {\em IEEE TPAMI}, 35(5):1080--1093, 2012.

\bibitem{Alldieck_2021_ICCV}
Thiemo Alldieck, Hongyi Xu, and Cristian Sminchisescu.
\newblock imghum: Implicit generative models of 3d human shape and articulated
  pose.
\newblock In {\em ICCV}, 2021.

\bibitem{amberg2007optimal}
Brian Amberg, Sami Romdhani, and Thomas Vetter.
\newblock Optimal step nonrigid icp algorithms for surface registration.
\newblock In {\em CVPR}, 2007.

\bibitem{bagautdinov2018modeling}
Timur Bagautdinov, Chenglei Wu, Jason Saragih, Pascal Fua, and Yaser Sheikh.
\newblock Modeling facial geometry using compositional vaes.
\newblock In {\em CVPR}, 2018.

\bibitem{bahri2021shape}
Mehdi Bahri, Eimear O’Sullivan, Shunwang Gong, Feng Liu, Xiaoming Liu,
  Michael~M Bronstein, and Stefanos Zafeiriou.
\newblock Shape my face: registering 3d face scans by surface-to-surface
  translation.
\newblock {\em IJCV}, 129(9):2680--2713, 2021.

\bibitem{blanz1999morphable}
Volker Blanz and Thomas Vetter.
\newblock A morphable model for the synthesis of 3d faces.
\newblock In {\em SIGGRAPH}, 1999.

\bibitem{blanz2003face}
Volker Blanz and Thomas Vetter.
\newblock Face recognition based on fitting a 3d morphable model.
\newblock {\em IEEE TPAMI}, 25(9):1063--1074, 2003.

\bibitem{bolkart2015groupwise}
Timo Bolkart and Stefanie Wuhrer.
\newblock A groupwise multilinear correspondence optimization for 3d faces.
\newblock In {\em ICCV}, 2015.

\bibitem{booth2018large}
James Booth, Anastasios Roussos, Allan Ponniah, David Dunaway, and Stefanos
  Zafeiriou.
\newblock Large scale 3d morphable models.
\newblock {\em IJCV}, 126(2):233--254, 2018.

\bibitem{bouritsas2019neural}
Giorgos Bouritsas, Sergiy Bokhnyak, Stylianos Ploumpis, Michael Bronstein, and
  Stefanos Zafeiriou.
\newblock Neural 3d morphable models: Spiral convolutional networks for 3d
  shape representation learning and generation.
\newblock In {\em ICCV}, 2019.

\bibitem{brunton2014multilinear}
Alan Brunton, Timo Bolkart, and Stefanie Wuhrer.
\newblock Multilinear wavelets: A statistical shape space for human faces.
\newblock In {\em ECCV}, 2014.

\bibitem{cao2013facewarehouse}
Chen Cao, Yanlin Weng, Shun Zhou, Yiying Tong, and Kun Zhou.
\newblock Facewarehouse: A 3d facial expression database for visual computing.
\newblock {\em IEEE TVCG}, 20(3):413--425, 2013.

\bibitem{Chen_2021_ICCV}
Xu Chen, Yufeng Zheng, Michael~J. Black, Otmar Hilliges, and Andreas Geiger.
\newblock Snarf: Differentiable forward skinning for animating non-rigid neural
  implicit shapes.
\newblock In {\em ICCV}, 2021.

\bibitem{chen2021learning}
Zhixiang Chen and Tae-Kyun Kim.
\newblock Learning feature aggregation for deep 3d morphable models.
\newblock In {\em CVPR}, 2021.

\bibitem{chen2019learning}
Zhiqin Chen and Hao Zhang.
\newblock Learning implicit fields for generative shape modeling.
\newblock In {\em CVPR}, 2019.

\bibitem{cheng2019arxiv}
Shiyang Cheng, Michael~M. Bronstein, Yuxiang Zhou, Irene Kotsia, Maja Pantic,
  and Stefanos Zafeiriou.
\newblock Meshgan: Non-linear 3d morphable models of faces.
\newblock {\em CoRR}, abs/1903.10384, 2019.

\bibitem{chibane2020ndf}
Julian Chibane, Aymen Mir, and Gerard Pons-Moll.
\newblock Neural unsigned distance fields for implicit function learning.
\newblock In {\em NeurIPS}, 2020.

\bibitem{corona2021smplicit}
Enric Corona, Albert Pumarola, Guillem Alenya, Gerard Pons-Moll, and Francesc
  Moreno-Noguer.
\newblock Smplicit: Topology-aware generative model for clothed people.
\newblock In {\em CVPR}, 2021.

\bibitem{cosker2011facs}
Darren Cosker, Eva Krumhuber, and Adrian Hilton.
\newblock A facs valid 3d dynamic action unit database with applications to 3d
  dynamic morphable facial modeling.
\newblock In {\em ICCV}, 2011.

\bibitem{deng2021deformed}
Yu Deng, Jiaolong Yang, and Xin Tong.
\newblock Deformed implicit field: Modeling 3d shapes with learned dense
  correspondence.
\newblock In {\em CVPR}, 2021.

\bibitem{egger20203d}
Bernhard Egger, William~AP Smith, Ayush Tewari, Stefanie Wuhrer, Michael
  Zollhoefer, Thabo Beeler, Florian Bernard, Timo Bolkart, Adam Kortylewski,
  Sami Romdhani, et~al.
\newblock 3d morphable face models—past, present, and future.
\newblock {\em ACM TOG}, 39(5):1--38, 2020.

\bibitem{genova2020local}
Kyle Genova, Forrester Cole, Avneesh Sud, Aaron Sarna, and Thomas Funkhouser.
\newblock Local deep implicit functions for 3d shape.
\newblock In {\em CVPR}, 2020.

\bibitem{genova2019iccv}
Kyle Genova, Forrester Cole, Daniel Vlasic, Aaron Sarna, William~T Freeman, and
  Thomas Funkhouser.
\newblock Learning shape templates with structured implicit functions.
\newblock In {\em ICCV}, 2019.

\bibitem{gilani2017dense}
Syed~Zulqarnain Gilani, Ajmal Mian, Faisal Shafait, and Ian Reid.
\newblock Dense 3d face correspondence.
\newblock {\em IEEE TPAMI}, 40(7):1584--1598, 2017.

\bibitem{icml2020_2086}
Amos Gropp, Lior Yariv, Niv Haim, Matan Atzmon, and Yaron Lipman.
\newblock Implicit geometric regularization for learning shapes.
\newblock In {\em ICML}. 2020.

\bibitem{hu2016face}
Guosheng Hu, Fei Yan, Chi-Ho Chan, Weihong Deng, William Christmas, Josef
  Kittler, and Neil~M Robertson.
\newblock Face recognition using a unified 3d morphable model.
\newblock In {\em ECCV}, 2016.

\bibitem{ibing20213d}
Moritz Ibing, Isaak Lim, and Leif Kobbelt.
\newblock 3d shape generation with grid-based implicit functions.
\newblock In {\em CVPR}, 2021.

\bibitem{lee1980two}
Der-Tsai Lee and Bruce~J Schachter.
\newblock Two algorithms for constructing a delaunay triangulation.
\newblock {\em International Journal of Computer \& Information Sciences},
  9(3):219--242, 1980.

\bibitem{lewiner2003efficient}
Thomas Lewiner, H{\'e}lio Lopes, Ant{\^o}nio~Wilson Vieira, and Geovan Tavares.
\newblock Efficient implementation of marching cubes' cases with topological
  guarantees.
\newblock {\em Journal of graphics tools}, 8(2):1--15, 2003.

\bibitem{lewis2000pose}
John~P Lewis, Matt Cordner, and Nickson Fong.
\newblock Pose space deformation: a unified approach to shape interpolation and
  skeleton-driven deformation.
\newblock In {\em SIGGRAPH}, 2000.

\bibitem{li2017flame}
Tianye Li, Timo Bolkart, Michael~J Black, Hao Li, and Javier Romero.
\newblock Learning a model of facial shape and expression from 4d scans.
\newblock {\em ACM TOG}, 36(6):194--1, 2017.

\bibitem{pmlr-v139-lipman21a}
Yaron Lipman.
\newblock Phase transitions, distance functions, and implicit neural
  representations.
\newblock In {\em ICML}, 2021.

\bibitem{liu2020nips}
Feng Liu and Xiaoming Liu.
\newblock Learning implicit functions for topology-varying dense 3d shape
  correspondence.
\newblock In {\em NeurIPS}, 2020.

\bibitem{liu20193d}
Feng Liu, Luan Tran, and Xiaoming Liu.
\newblock 3d face modeling from diverse raw scan data.
\newblock In {\em ICCV}, 2019.

\bibitem{mescheder2019occupancy}
Lars Mescheder, Michael Oechsle, Michael Niemeyer, Sebastian Nowozin, and
  Andreas Geiger.
\newblock Occupancy networks: Learning 3d reconstruction in function space.
\newblock In {\em CVPR}, 2019.

\bibitem{mildenhall2020nerf}
Ben Mildenhall, Pratul~P Srinivasan, Matthew Tancik, Jonathan~T Barron, Ravi
  Ramamoorthi, and Ren Ng.
\newblock Nerf: Representing scenes as neural radiance fields for view
  synthesis.
\newblock In {\em ECCV}, 2020.

\bibitem{moller1997fast}
Tomas M{\"o}ller and Ben Trumbore.
\newblock Fast, minimum storage ray-triangle intersection.
\newblock {\em Journal of graphics tools}, 2(1):21--28, 1997.

\bibitem{park2019deepsdf}
Jeong~Joon Park, Peter Florence, Julian Straub, Richard Newcombe, and Steven
  Lovegrove.
\newblock Deepsdf: Learning continuous signed distance functions for shape
  representation.
\newblock In {\em CVPR}, 2019.

\bibitem{patel20093d}
Ankur Patel and William~AP Smith.
\newblock 3d morphable face models revisited.
\newblock In {\em CVPR}, 2009.

\bibitem{bfm09}
Pascal Paysan, Reinhard Knothe, Brian Amberg, Sami Romdhani, and Thomas Vetter.
\newblock A 3d face model for pose and illumination invariant face recognition.
\newblock In {\em AVSS}, 2009.

\bibitem{peng2021animatable}
Sida Peng, Junting Dong, Qianqian Wang, Shangzhan Zhang, Qing Shuai, Xiaowei
  Zhou, and Hujun Bao.
\newblock Animatable neural radiance fields for modeling dynamic human bodies.
\newblock In {\em ICCV}, 2021.

\bibitem{peng2020convolutional}
Songyou Peng, Michael Niemeyer, Lars Mescheder, Marc Pollefeys, and Andreas
  Geiger.
\newblock Convolutional occupancy networks.
\newblock In {\em ECCV}, 2020.

\bibitem{peng2021neural}
Sida Peng, Yuanqing Zhang, Yinghao Xu, Qianqian Wang, Qing Shuai, Hujun Bao,
  and Xiaowei Zhou.
\newblock Neural body: Implicit neural representations with structured latent
  codes for novel view synthesis of dynamic humans.
\newblock In {\em CVPR}, 2021.

\bibitem{R_2021_CVPR}
Mallikarjun~B R, Ayush Tewari, Hans-Peter Seidel, Mohamed Elgharib, and
  Christian Theobalt.
\newblock Learning complete 3d morphable face models from images and videos.
\newblock In {\em CVPR}, 2021.

\bibitem{ramon2021h3d}
Eduard Ramon, Gil Triginer, Janna Escur, Albert Pumarola, Jaime Garcia, Xavier
  Giro-i Nieto, and Francesc Moreno-Noguer.
\newblock H3d-net: Few-shot high-fidelity 3d head reconstruction.
\newblock In {\em ICCV}, 2021.

\bibitem{ranjan2018generating}
Anurag Ranjan, Timo Bolkart, Soubhik Sanyal, and Michael~J Black.
\newblock Generating 3d faces using convolutional mesh autoencoders.
\newblock In {\em ECCV}, 2018.

\bibitem{saito2019pifu}
Shunsuke Saito, Zeng Huang, Ryota Natsume, Shigeo Morishima, Angjoo Kanazawa,
  and Hao Li.
\newblock Pifu: Pixel-aligned implicit function for high-resolution clothed
  human digitization.
\newblock In {\em ICCV}, 2019.

\bibitem{saito2020pifuhd}
Shunsuke Saito, Tomas Simon, Jason Saragih, and Hanbyul Joo.
\newblock Pifuhd: Multi-level pixel-aligned implicit function for
  high-resolution 3d human digitization.
\newblock In {\em CVPR}, 2020.

\bibitem{sitzmann2019metasdf}
Vincent Sitzmann, Eric~R. Chan, Richard Tucker, Noah Snavely, and Gordon
  Wetzstein.
\newblock Metasdf: Meta-learning signed distance functions.
\newblock In {\em NeurIPS}, 2020.

\bibitem{sitzmann2019siren}
Vincent Sitzmann, Julien~N.P. Martel, Alexander~W. Bergman, David~B. Lindell,
  and Gordon Wetzstein.
\newblock Implicit neural representations with periodic activation functions.
\newblock In {\em NeurIPS}, 2020.

\bibitem{staal2015describing}
Femke~CR Staal, Allan~JT Ponniah, Freida Angullia, Clifford Ruff, Maarten~J
  Koudstaal, and David Dunaway.
\newblock Describing crouzon and pfeiffer syndrome based on principal component
  analysis.
\newblock {\em Journal of Cranio-Maxillofacial Surgery}, 43(4):528--536, 2015.

\bibitem{takikawa2021neural}
Towaki Takikawa, Joey Litalien, Kangxue Yin, Karsten Kreis, Charles Loop, Derek
  Nowrouzezahrai, Alec Jacobson, Morgan McGuire, and Sanja Fidler.
\newblock Neural geometric level of detail: Real-time rendering with implicit
  3d shapes.
\newblock In {\em CVPR}, 2021.

\bibitem{tang2021octfield}
Jia-Heng Tang, Weikai Chen, Jie Yang, Bo Wang, Songrun Liu, Bo Yang, and Lin
  Gao.
\newblock Octfield: Hierarchical implicit functions for 3d modeling.
\newblock In {\em NeurIPS}, 2021.

\bibitem{tran2019towards}
Luan Tran, Feng Liu, and Xiaoming Liu.
\newblock Towards high-fidelity nonlinear 3d face morphable model.
\newblock In {\em CVPR}, 2019.

\bibitem{tran2018nonlinear}
Luan Tran and Xiaoming Liu.
\newblock Nonlinear 3d face morphable model.
\newblock In {\em CVPR}, 2018.

\bibitem{vandermaaten08a}
Laurens van~der Maaten and Geoffrey Hinton.
\newblock Visualizing data using t-sne.
\newblock {\em Journal of Machine Learning Research}, 9(86):2579--2605, 2008.

\bibitem{vlasic2006face}
Daniel Vlasic, Matthew Brand, Hanspeter Pfister, and Jovan Popovic.
\newblock Face transfer with multilinear models.
\newblock In {\em ACM SIGGRAPH Courses}. 2006.

\bibitem{wang2021neus}
Peng Wang, Lingjie Liu, Yuan Liu, Christian Theobalt, Taku Komura, and Wenping
  Wang.
\newblock Neus: Learning neural implicit surfaces by volume rendering for
  multi-view reconstruction.
\newblock In {\em NeurIPS}, 2021.

\bibitem{xu2019disn}
Qiangeng Xu, Weiyue Wang, Duygu Ceylan, Radomir Mech, and Ulrich Neumann.
\newblock Disn: Deep implicit surface network for high-quality single-view 3d
  reconstruction.
\newblock In {\em NeurIPS}, 2019.

\bibitem{yang2020facescape}
Haotian Yang, Hao Zhu, Yanru Wang, Mingkai Huang, Qiu Shen, Ruigang Yang, and
  Xun Cao.
\newblock Facescape: a large-scale high quality 3d face dataset and detailed
  riggable 3d face prediction.
\newblock In {\em CVPR}, 2020.

\bibitem{yenamandra2021i3dmm}
Tarun Yenamandra, Ayush Tewari, Florian Bernard, Hans-Peter Seidel, Mohamed
  Elgharib, Daniel Cremers, and Christian Theobalt.
\newblock i3dmm: Deep implicit 3d morphable model of human heads.
\newblock In {\em CVPR}, 2021.

\bibitem{yi2016scalable}
Li Yi, Vladimir~G Kim, Duygu Ceylan, I-Chao Shen, Mengyan Yan, Hao Su, Cewu Lu,
  Qixing Huang, Alla Sheffer, and Leonidas Guibas.
\newblock A scalable active framework for region annotation in 3d shape
  collections.
\newblock {\em ACM TOG}, 35(6):1--12, 2016.

\bibitem{yin2006bu3d}
Lijun Yin, Xiaozhou Wei, Yi Sun, Jun Wang, and Matthew~J Rosato.
\newblock A 3d facial expression database for facial behavior research.
\newblock In {\em FGR}, 2006.

\bibitem{zhang2021learning}
Jingyang Zhang, Yao Yao, and Long Quan.
\newblock Learning signed distance field for multi-view surface reconstruction.
\newblock In {\em ICCV}, 2021.

\bibitem{zheng2021deep}
Zerong Zheng, Tao Yu, Qionghai Dai, and Yebin Liu.
\newblock Deep implicit templates for 3d shape representation.
\newblock In {\em CVPR}, 2021.

\end{thebibliography}
}

\clearpage
\appendix  
\setcounter{table}{0}  
\setcounter{figure}{0}
\renewcommand\thesection{\Alph{section}}
\renewcommand{\thetable}{A\arabic{table}}
\renewcommand{\thefigure}{A\arabic{figure}}

\begin{strip}%
  \centering
  \large
  \textbf{%
  {ImFace: A Nonlinear 3D Morphable Face Model with Implicit Neural Representations}\\
  \vspace{0.3cm} \textit{Supplementary material} \\ \vspace{0.5cm}
  }
  \includegraphics[width=1\linewidth]{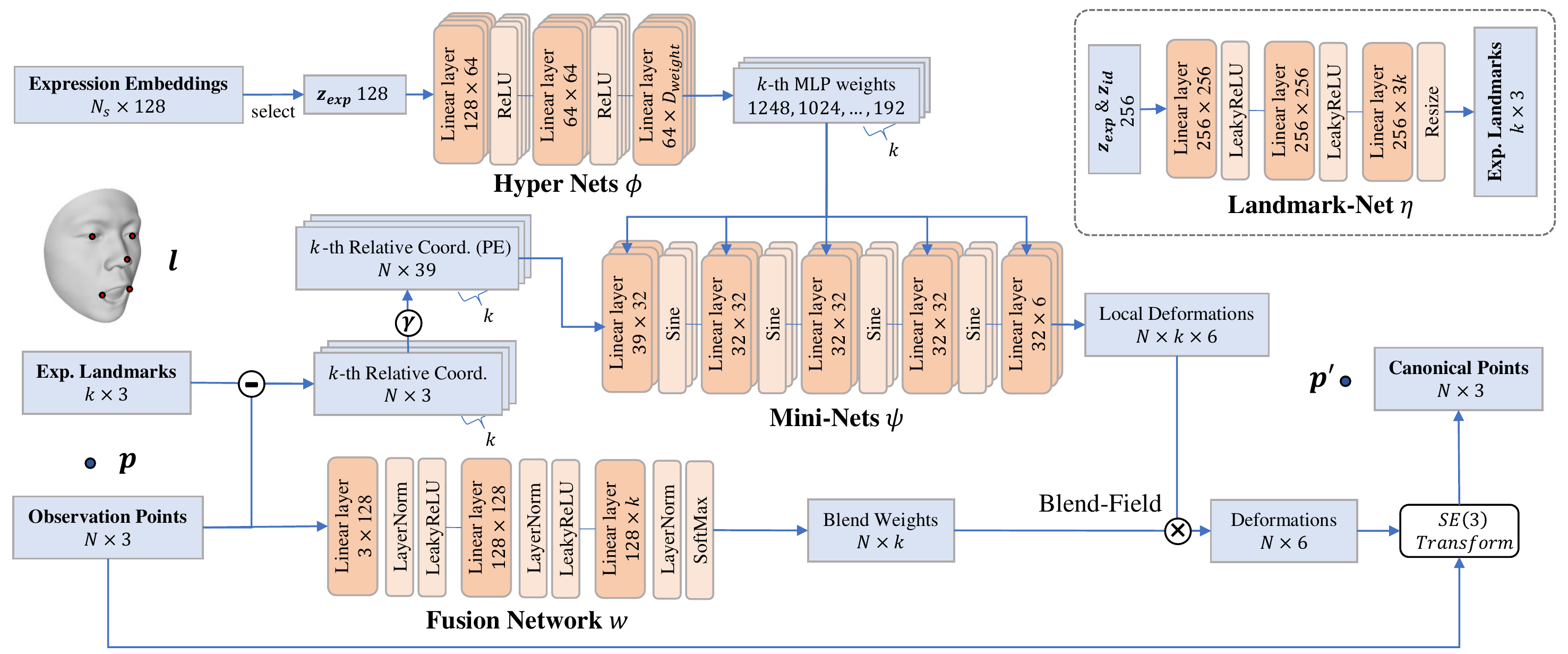}
  \captionof{figure}{Detailed architecture of Expression Mini-Nets block.}
  \label{fig:supp_net_D}
\end{strip}

\section{Detailed Network Architecture}
\begin{figure*}[!ht]
  \centering
  \includegraphics[width=1\linewidth]{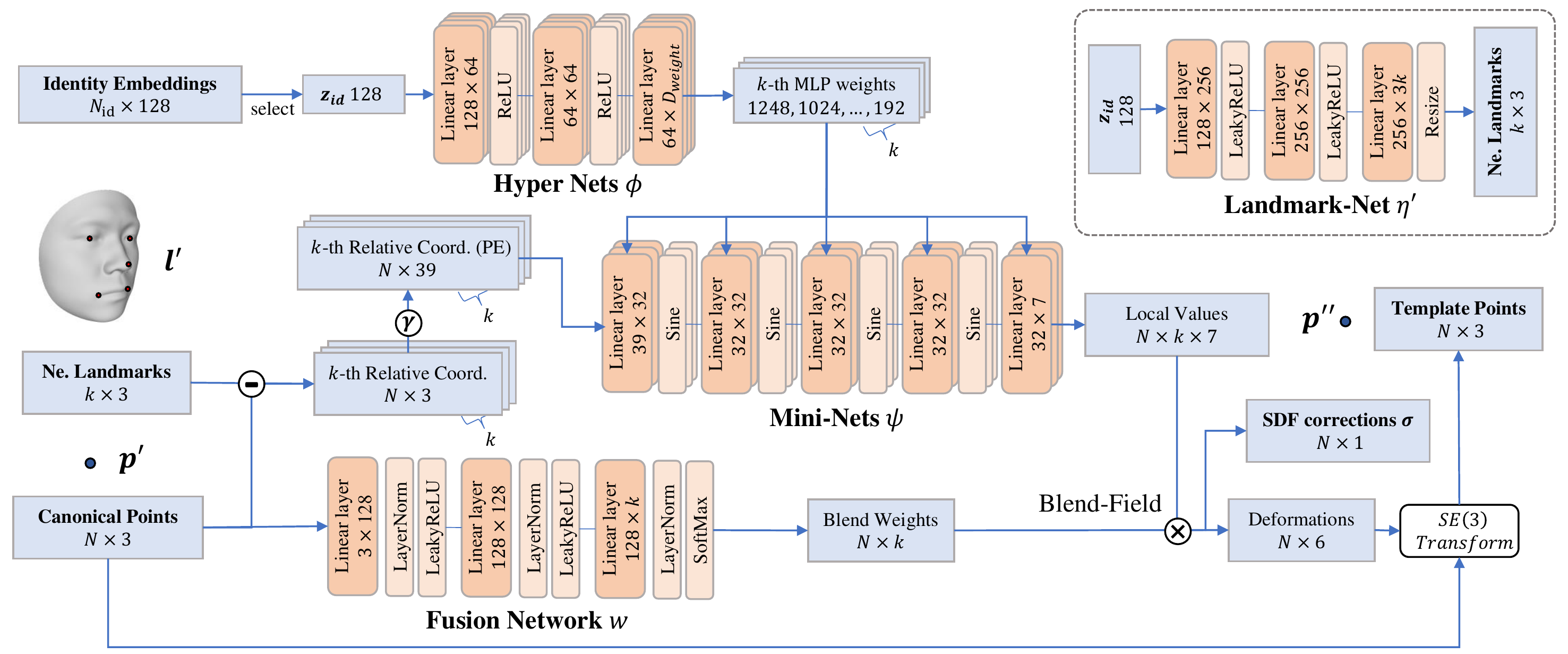}
  \caption{Detailed architecture of Identity Mini-Nets block. It additionally predicts a correction term to cope with possible non-existent correspondences.}
  \label{fig:supp_net_R}
\end{figure*}

\begin{figure*}[!ht]
  \centering
  \includegraphics[width=1\linewidth]{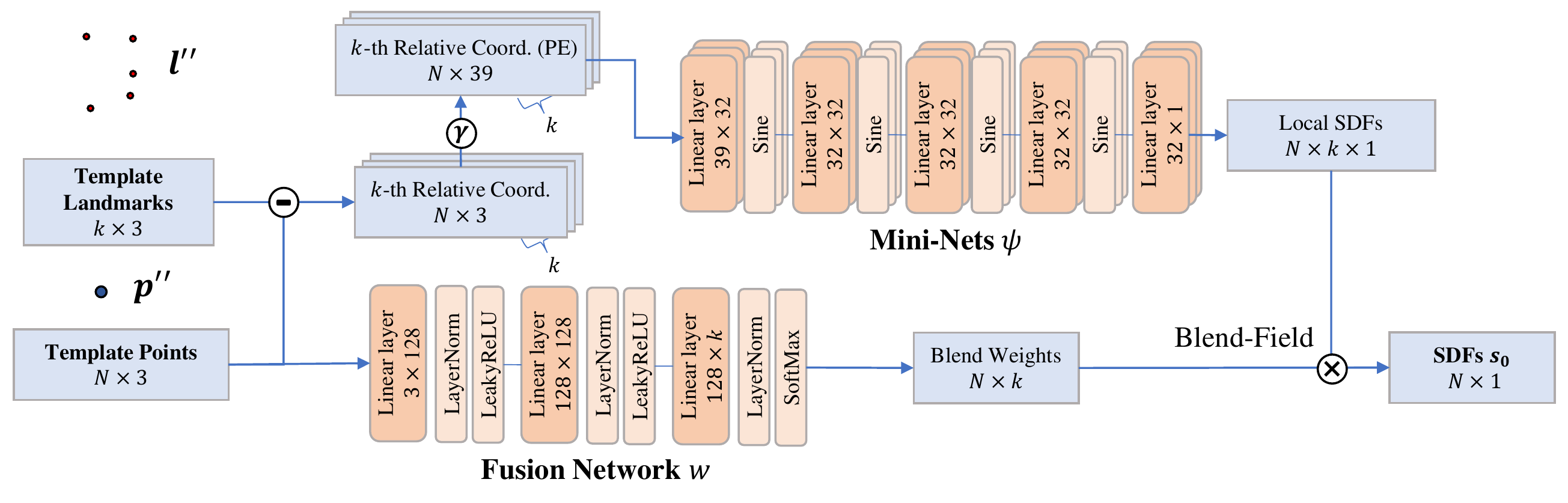}
  \caption{Detailed architecture of Template Mini-Nets block. It does not require a conditional embedding since the template face is shared across all faces.}
  \label{fig:supp_net_T}
\end{figure*}

 The detailed architectures of Expression Mini-Nets, Identity Mini-Nets, and Template Mini-Nets are shown in Fig.~\ref{fig:supp_net_D}, Fig.~\ref{fig:supp_net_R} and Fig.~\ref{fig:supp_net_T} respectively, where $N_s$ refers to the number of total scans, $N_{id}$ denotes the number of identities, and PE indicates positional encoding.
 
 All the networks are fully implemented by MLPs. To achieve better performance for high-frequency clues, we encode the relative coordinates with respect to $k$ landmarks by sinusoidal positional encoding $\gamma$\cite{mildenhall2020nerf}, written as $\gamma(p)\!=\!(\sin(2^0\pi p),\cos(2^0\pi p),\cdots,\sin(2^{L-1}\pi p),\cos(2^{L-1}\pi p))$, where $L\!=\!6$, $k\!=\!5$ in our experiments. Besides, sine activations are exploited in every Mini-Net $\psi_n$ and the parameters are initialized as in \cite{sitzmann2019siren}. During training, the parameters of $k$ Mini-Nets are generated by $k$ corresponding Hyper Nets for a more expressive latent space, which is a common technical operation in recent INRs studies \cite{deng2021deformed,sitzmann2019metasdf}.

 The trade-off parameters $\lambda_1,\cdots,\lambda_7$ to train the networks are set to $3e3$, $1e2$, $5e1$, $1e6$, $1e2$, $1e2$, and $1e2$ respectively. 

\begin{figure}
  \centering
  \setlength{\abovecaptionskip}{0pt}
  \setlength{\belowcaptionskip}{0pt}
  \includegraphics[width=1\linewidth]{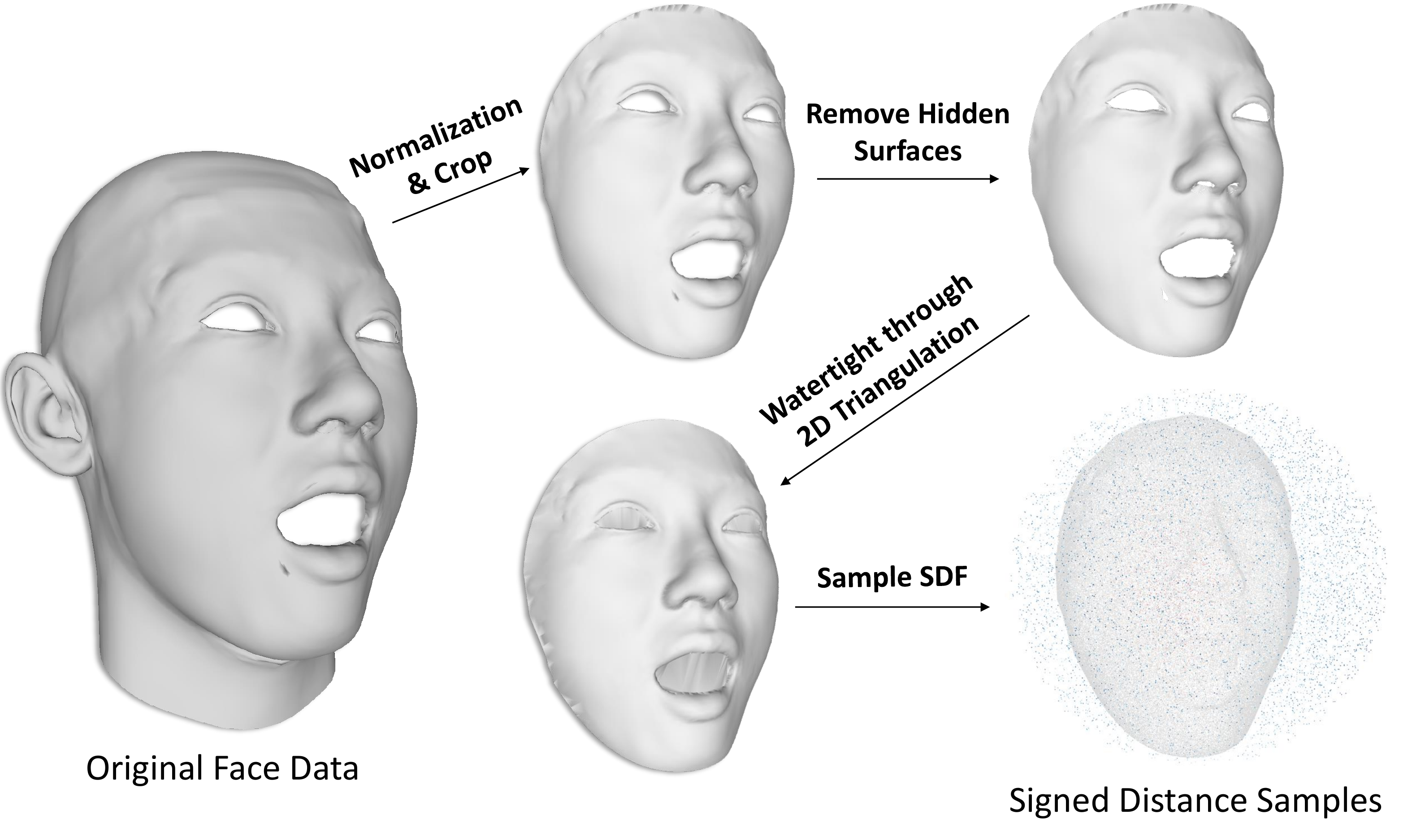}
  \caption{Illustration of the preprocessing pipeline.}
  % \vspace{-4mm}
  \label{fig:supp_pre}
\end{figure}

\section{Preprocessing}
\label{sec:Sampling}

\begin{figure*}
  \centering
  \includegraphics[width=1\linewidth]{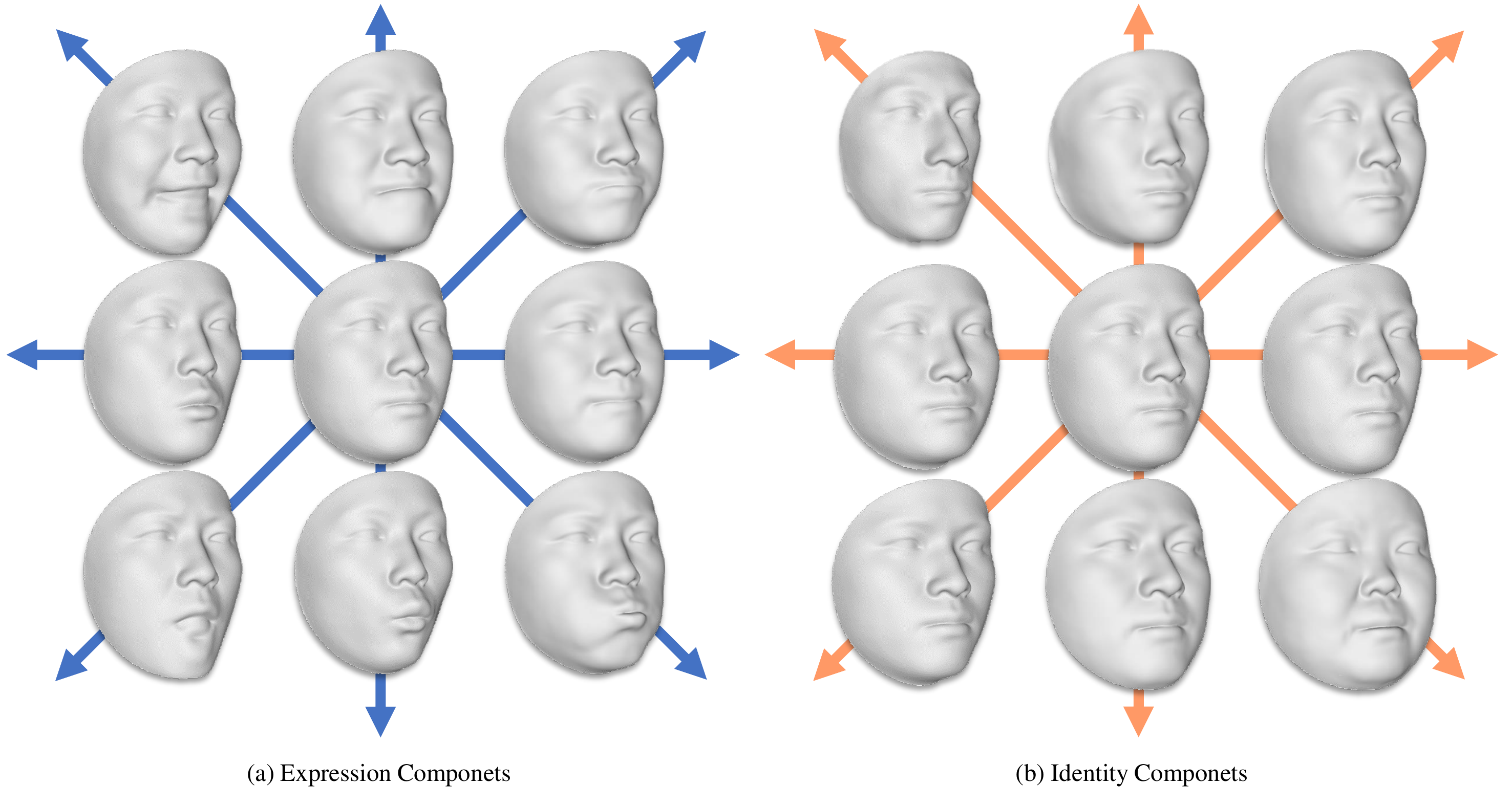}
  % \vspace{-5mm}
  \caption{Visualization of the expression and identity principal components of ImFace on FaceScape \cite{yang2020facescape}. }
  \label{fig:pca}
\end{figure*}

To enable INRs to work with non-watertight 3D faces, we present an effective preprocessing pipeline as briefly described in Sec.~3.5, so that facial geometry and correspondence can be learned as exquisitely as on  watertight objects. 

We first determine the domain of definition $\{(x,y,z)\in{\mathbb R}^3\}$ of our implicit function $f: {\mathbb R}^3 \mapsto {\mathbb R}$. Specifically, the coordinate origin is set at the point 4 $cm$ behind the nose tip. This setting helps to balance the number of positive and negative SDF samples, which is crucial to facilitate INRs network convergence. To cover most of facial geometries while cut away unnecessary regions, a sphere $S$ with a radius of 10 $cm$ centered on the coordinate origin is defined as the sampling area. 
However, it is not an intuitive task to determine whether a point in $S$ is ``inside" or ``outside" of a 3D facial surface, mainly because facial surface may contain multiple openings such as the mouth and eyes, as well as complex geometric structures in the nasal or oral cavity, depending on the acquisition conditions. The proposed preprocessing pipeline aims to address the issue above.

Our key observation is that, if a 3D surface $C$ satisfies the following property, it can be oriented easily. Without loss of generality, we consider an infinite continuous surface $C$ defined in 3D space $\Omega:\{(x,y,z)\in{\mathbb R}^3\}$.

\noindent\emph{\textbf{Definition 1.} Assuming $C$ is defined by an implicit function $f(x,y,z)=0$, if $z$ is an injective function of $(x,y)$, we call $C$ is injective at $z$.
}

\noindent\emph{\textbf{Property 1.} If $C$ is injective at $z$, $\Omega$ is divided into two and only two spaces $\Omega_1$, $\Omega_2$, that for any $(x,y,z)\in\Omega_1$, a ray from $(x,y,z)$ along the positive z-axis intersects with $C$ for only once, and for any $(x,y,z)\in\Omega_2$ it does not intersect with $C$.
}

\noindent\emph{Proof.} It is equivalent to prove that for any $(x,y,z)\in\Omega$, the ray starting from it along the positive z-axis cannot intersect with $C$ at more than one point. By reductio, if a ray $l$ from $(x_0,y_0,z_0)\in\Omega$ intersects with $C$ at multiple different points $\{(x_1,y_1,z_1),\cdots,(x_n,y_n,z_n)\} (n\ge2)$, we have $(x_1,y_1)\!=\!\cdots\!=\!(x_n,y_n)\!=\!(x_0,y_0)$ but $z_1\!\ne\!\cdots\!\ne\!z_n$, which conflicts with the injective precondition of $z$.

Based on the property above, we acknowledge that if a facial surface satisfies the property in \emph{\textbf{Definition 1}}, then two separate 3D regions can be clearly determined and the inside and outside space can be set manually, so that SDF can be further defined. In general, human faces are approximately injective at the frontal direction. To make it strictly satisfy \emph{\textbf{Property 1}}, for any face mesh $(V,F)$ we generate new triangles $F'$ by performing the Delaunay Triangulation Algorithm\cite{lee1980two} on x-y coordinate and construct a new mesh $(V,F')$, thus any straight line parallel to z-axis only intersects with one triangle in $F'$. It holds because a Triangulation Algorithm covers the convex hull and does not lead to overlaps between 2D triangles, which makes any x-y coordinate have a unique triangle corresponding to it.

Considering that directly performing 2D triangulation makes the points on hidden surfaces (\textit{e.g.} the inner surface of nasal cavity) interlace with the ones on frontal surface, leading to unreasonable triangles, the Ray-Triangle Intersection Algorithm\cite{moller1997fast} is thus iteratively executed to remove the hidden surfaces before triangulation. Specifically, a vertex is marked if a ray from it along the positive z-axis intersects with more than one triangle in $F$, and the triangles in $F$ which have marked vertices are removed. In this way, a pseudo watertight face mesh can be established without much loss of accuracy, which divides the sampling space into two separate parts clearly.

Given a preprocessed face mesh, the sign of any query point in the sampling sphere can be determined by whether a ray from it to positive z-axis intersects with the face mesh, as in \emph{\textbf{Property 1}}. 
In order to accelerate the calculation procedure, we make use of the distance vectors calculated via distance transform and determine the sign of a query point by the angle between its distance vector to the nearest surface and the positive direction of z-axis , which is equivalent to the above-mentioned ray intersection checks. The whole preprocessing pipeline is presented in Fig.~\ref{fig:supp_pre}.

\begin{figure}
  \centering
  \setlength{\abovecaptionskip}{0pt}
  \setlength{\belowcaptionskip}{0pt}
  \begin{subfigure}{0.49\linewidth}
    \includegraphics[width=1\linewidth]{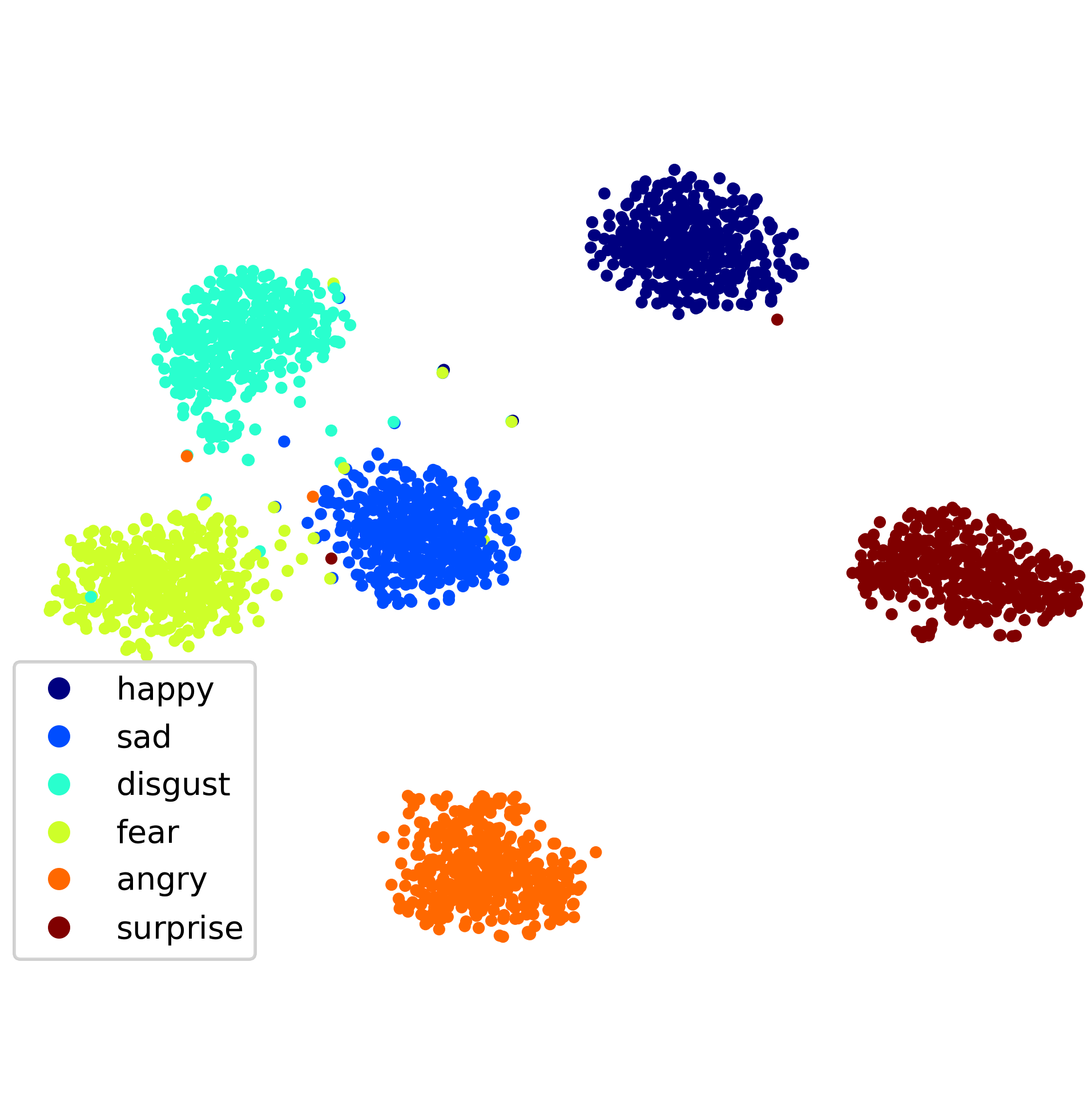}
    \vspace{-7mm}
    \caption{}
  \end{subfigure}
  \hfill
  \begin{subfigure}{0.49\linewidth}
    \includegraphics[width=1\linewidth]{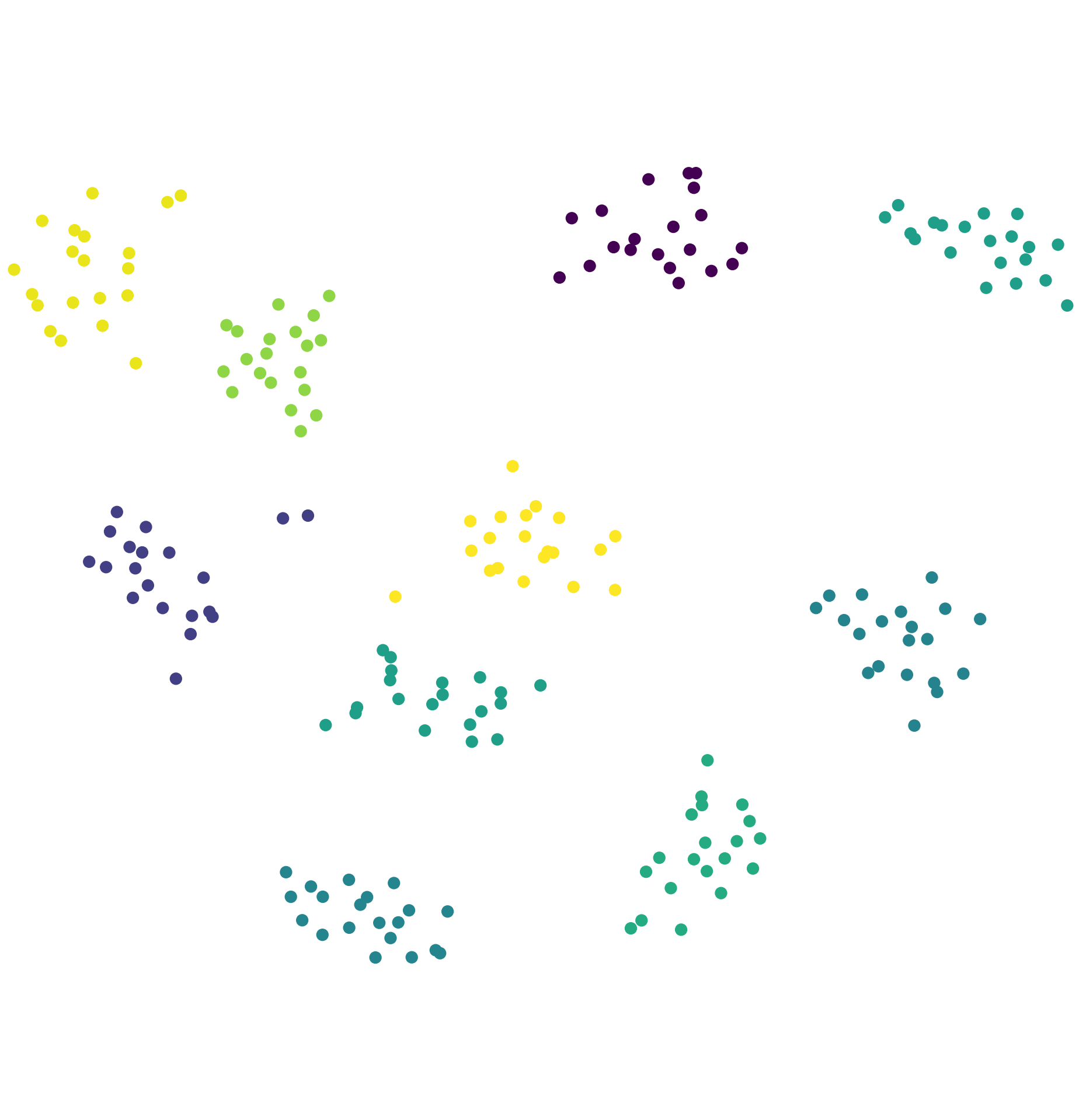}
    \vspace{-8mm}
    \caption{}
  \end{subfigure}
  \vspace{2mm}
  \caption{Visualizing the distributions of  high-dimensional expression and identity embeddings with t-SNE. \textbf{(a)} the embedding distribution of 6 typical expressions from the training set. \textbf{(b)} the identity embedding distribution from the test set. 
  }
  \label{fig:tsne}
\end{figure}

\begin{figure*}
  \centering
  \includegraphics[width=1\linewidth]{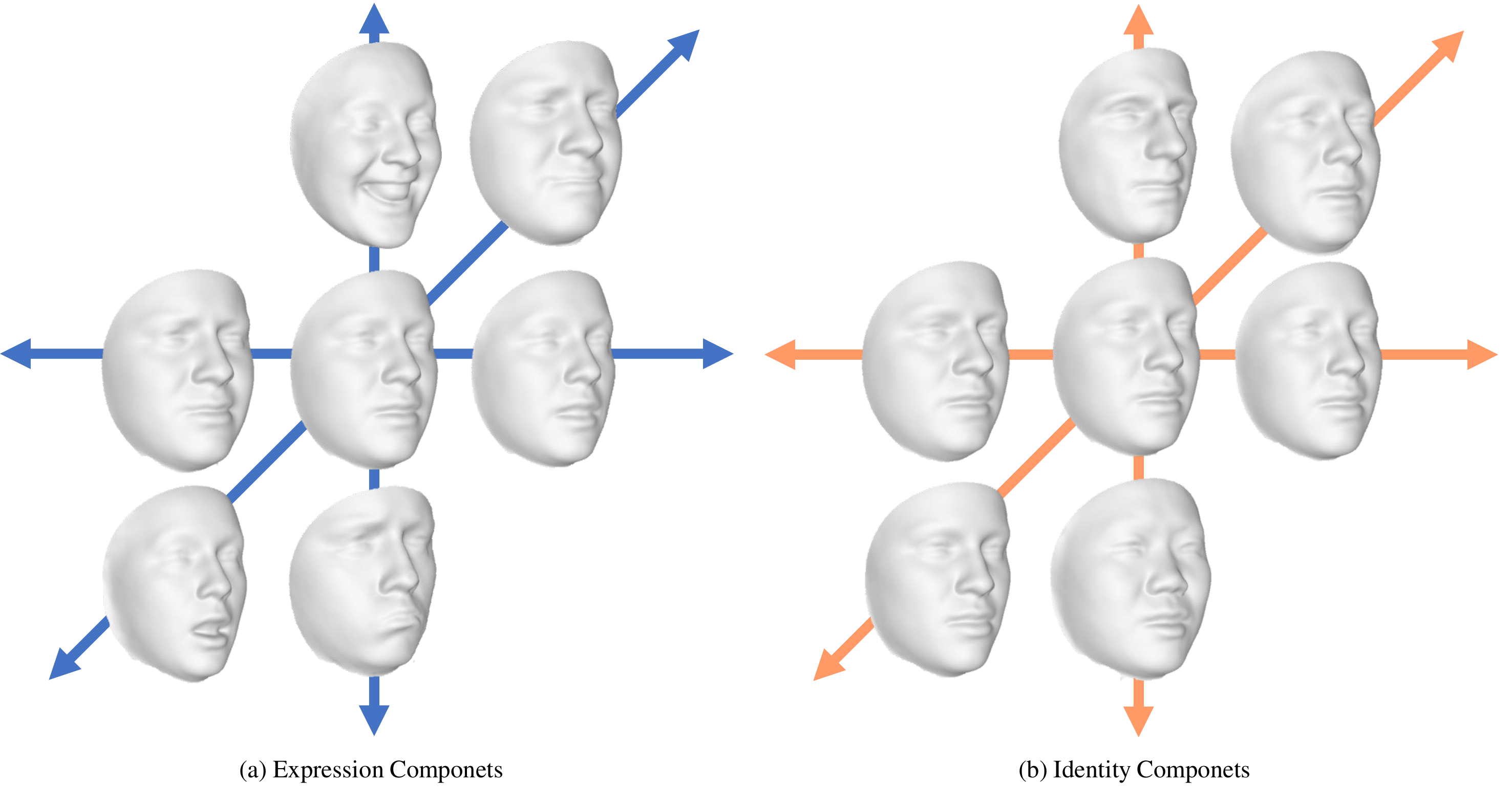}
  \vspace{-8mm}
  \caption{Visualization of the expression and identity principal components on BU-3DFE\cite{yin2006bu3d}.
  }
  \label{fig:BU3D_PCA}
\end{figure*}

\begin{figure}[t]
  \centering
  \includegraphics[width=0.95\linewidth]{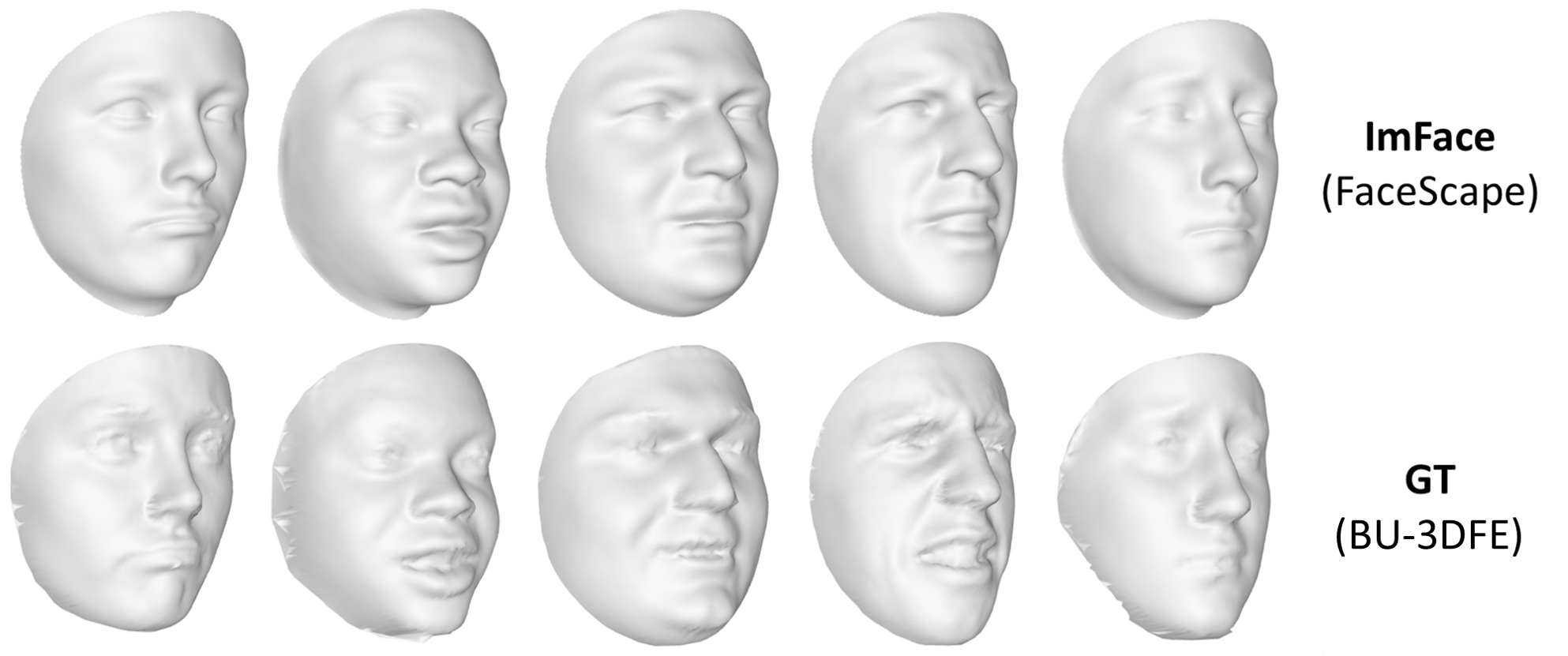}
   \caption{Cross-dataset reconstruction (trained on FaceScape, tested on BU-3DFE).}
   \label{fig:cross}
\end{figure}

\section{Experiments}

To have a better insight into the proposed ImFace morphable model, we provide more evaluation results in this supplementary material.

\subsection{Face Variation Visualization}

We apply Principal Components Analysis (PCA) on the learned expression and identity embeddings to visualize model variations, as shown in Fig.~\ref{fig:pca}. The standard deviations in terms of expression and identity are set to $\pm 3$ and $\pm 30$ respectively. In particular, four expression principal components are visualized in Fig.~\ref{fig:pca} (a). Despite great expression changes, the faces maintain a consistent identity. Besides, by learning expression components from thousands of unique embeddings, vivid expressions can be produced by ImFace. In Fig.~\ref{fig:pca} (b), we can observe a similar phenomenon on the learned identity components that the facial expressions remain stable when identity varies. The experimental results indicate that a good distanglement between expression and identity is achieved, which is crucial to generating novel faces by reweighting the singular values.

\subsection{High-dimensional Embedding Visualization}

To validate this point, we visualize the learned high-dimensional expression embeddings from 2,130 training scans with 6 typical expressions by t-SNE\cite{vandermaaten08a}, as  Fig.~\ref{fig:tsne} (a) shows. It can be seen that our network is capable of unsupervisedly distinguishing different expression types only by learning from expression-related shape morphs, which indicates its superior ability in expression modeling. Furthermore, we visualize the identity embeddings from the test set in Fig.~\ref{fig:tsne} (b), which involves 200 face scans from 10 persons. Visually inspected, our model successfully captures different identity features even under various complicated expressions.

\subsection{More Results on FaceScape}

\begin{figure*}
  \centering
  \includegraphics[width=0.94\linewidth]{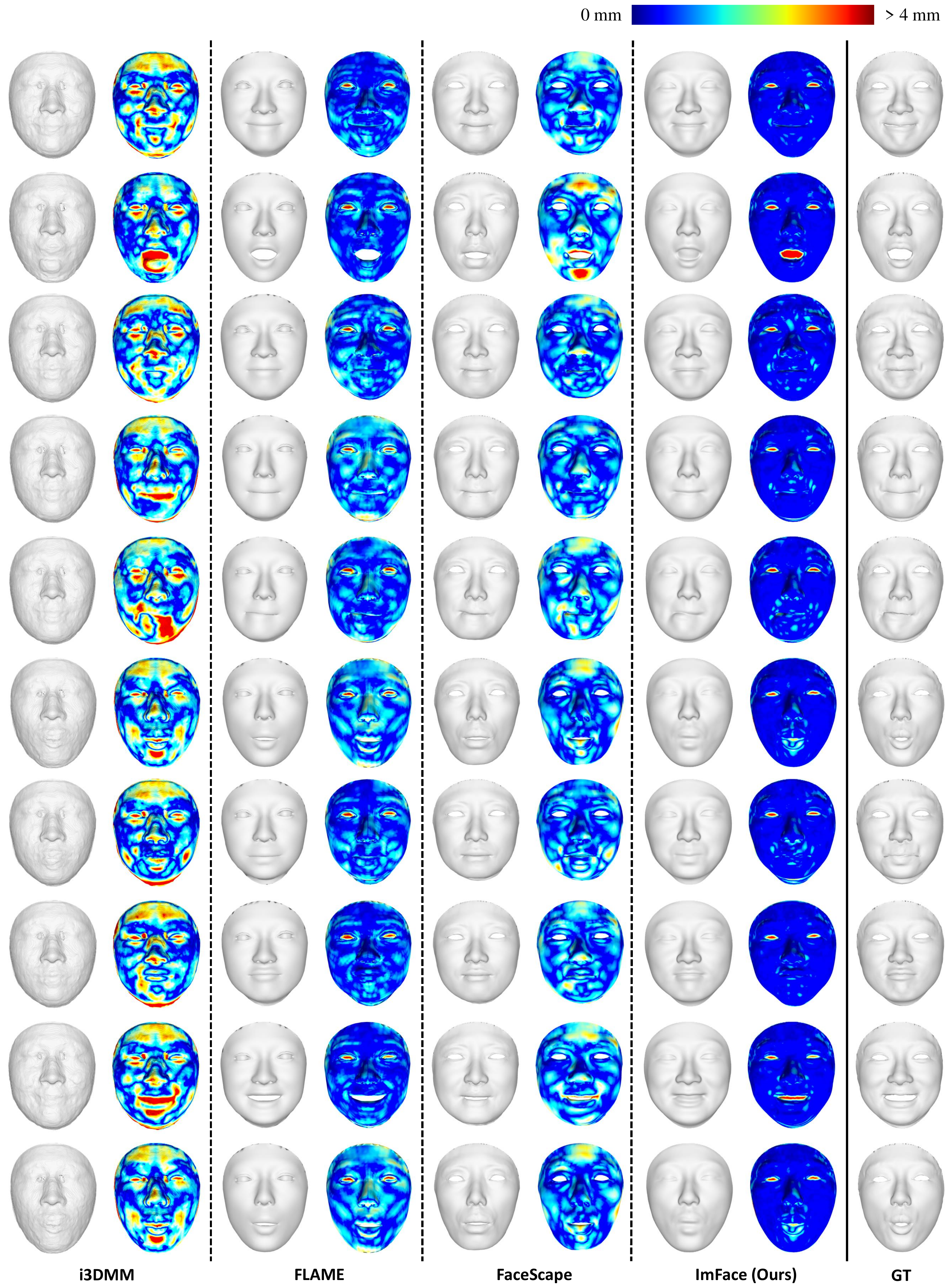}
  \vspace{-2mm}
  \caption{More comparison with i3DMM\cite{yenamandra2021i3dmm}, FLAME\cite{li2017flame}, FaceScape\cite{yang2020facescape}, and ground-truth faces.}
  \label{fig:supp_error}
\end{figure*}

\begin{figure*}
  \centering
  \includegraphics[width=0.94\linewidth]{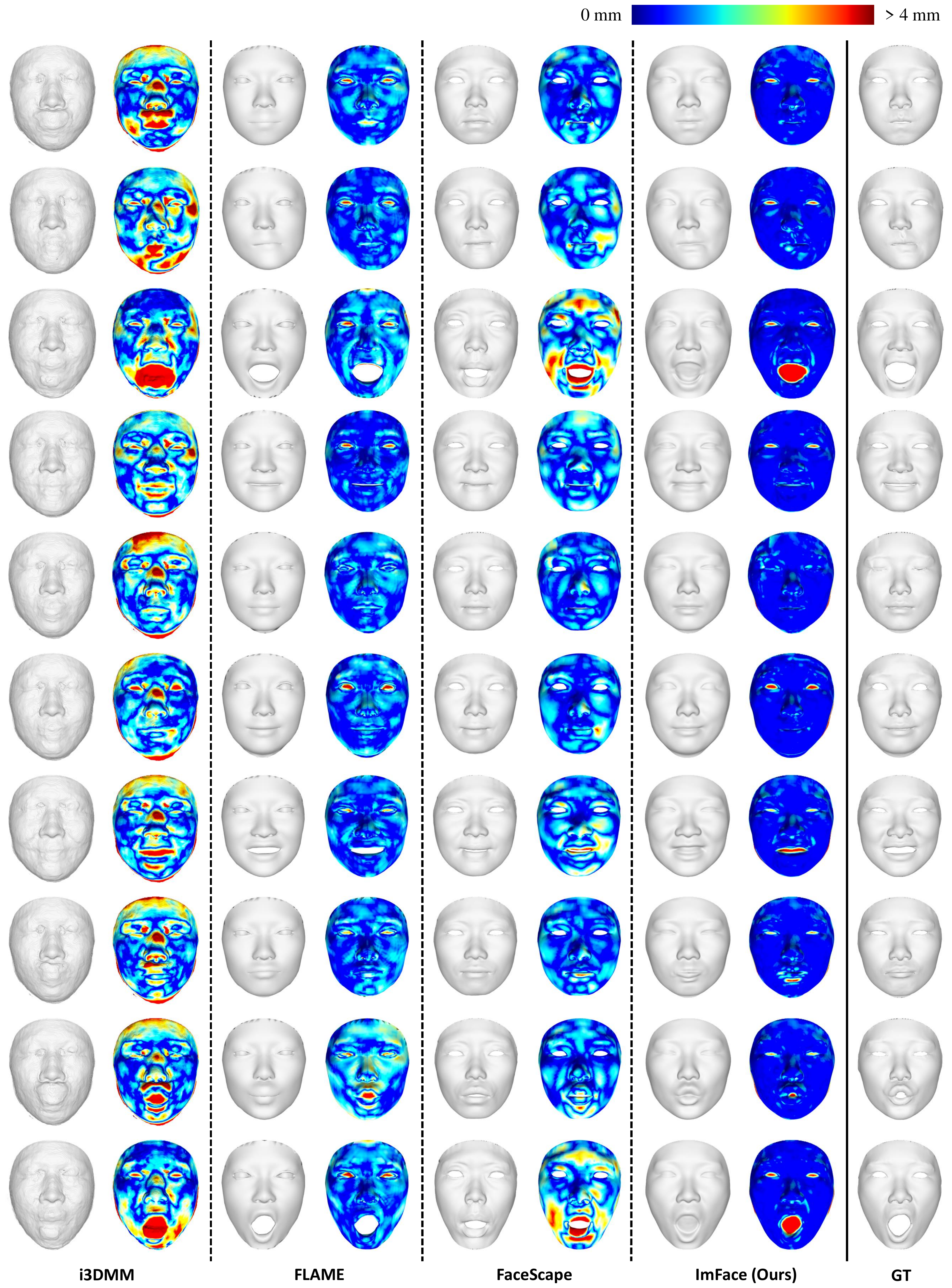}
  \vspace{-2mm}
  \caption{More comparison with i3DMM\cite{yenamandra2021i3dmm}, FLAME\cite{li2017flame}, FaceScape\cite{yang2020facescape}, and ground-truth faces.}
  \label{fig:supp_error2}
\end{figure*}

In Fig.~\ref{fig:supp_error} and Fig.~\ref{fig:supp_error2}, we present more comparison with i3DMM\cite{yenamandra2021i3dmm}, FLAME\cite{li2017flame}, FaceScape\cite{yang2020facescape}, and ground-truth faces, where a common color-coded distance (fit-to-scan) is used to indicate the reconstruction errors. As can be seen, faces are reconstructed more accurately by ImFace than the counterparts.

\subsection{Results on BU-3DFE}

We additionally preprocess the BU-3DFE\cite{yin2006bu3d}  database and train ImFace on it. Fig.~\ref{fig:BU3D_PCA} displays the expression and identity principal components achieved on BU-3DFE. As can be seen, although this dataset contains some pose variations, ImFace still captures facial geometry faithfully, which validates its generality. Further, we give cross-dataset results in Fig.~\ref{fig:cross}, which shows that our model well generalizes to another dataset.

\subsection{Applications}

\begin{figure*}
  \centering
  \includegraphics[width=1\linewidth]{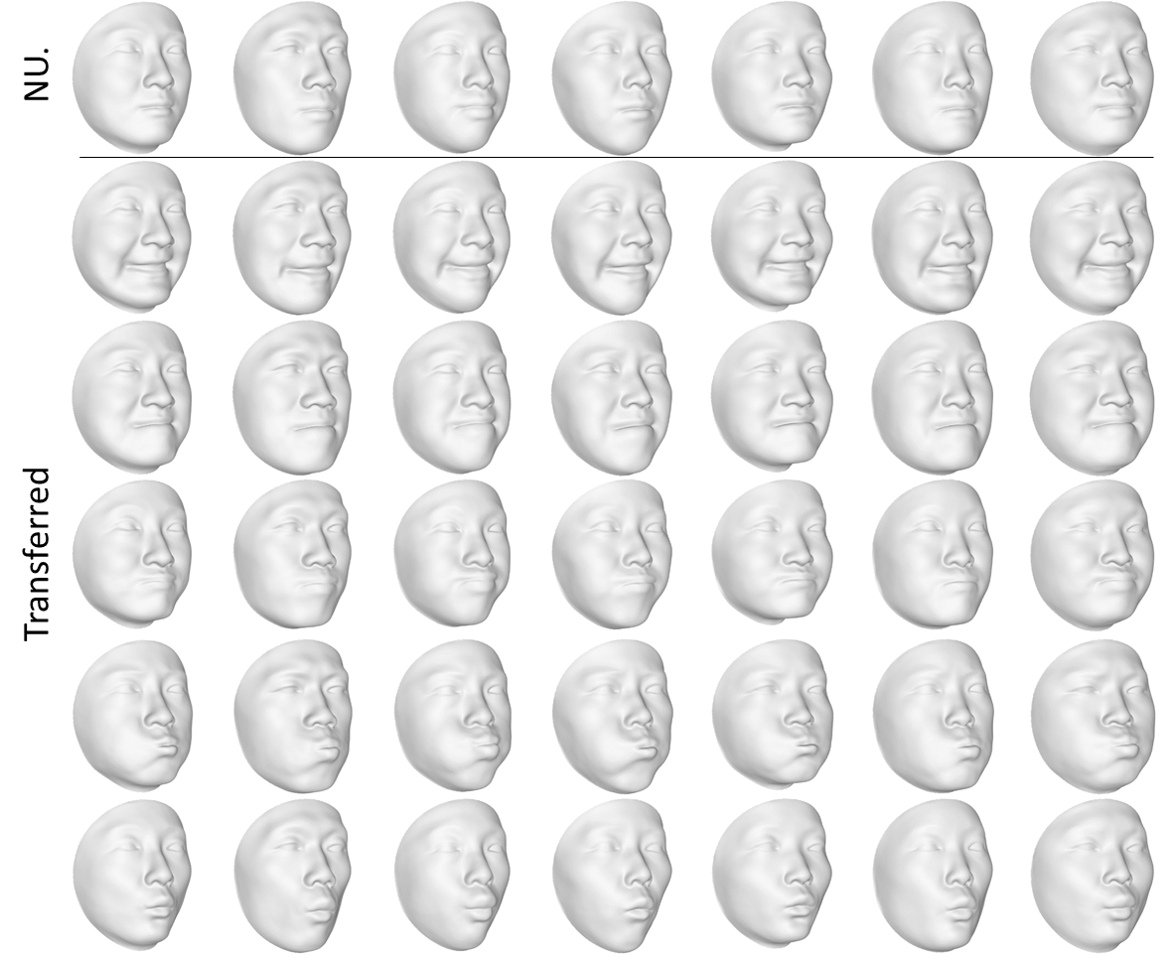}
  \vspace{-2mm}
  \caption{Expression editing results on the test set. The faces in the first row provide source identities, while the rest are generated by the proposed model. Our model is able to edit facial expression by exchanging embeddings.}
  \label{fig:supp_trans}
\end{figure*}

ImFace is a general face representation established upon prior distributions of facial expression and identity morphs, and it can thus  be applied to various down-stream applications. In Fig.~\ref{fig:supp_trans}, we provide expression editing results achieved on the FaceScape test set. The faces in the first row are the real ones providing source identities with neutral expressions, while the rest are generated by the proposed model. Our model is able to edit facial expression by simply changing their expression embeddings. The vivid 3D faces generated clearly validate the powerful representation ability of ImFace.

\section{Visualization Techniques}

We use Marching Cubes\cite{lewiner2003efficient} to reconstruct facial surfaces from the signed distance field, where the voxel resolution is set to $256^3$. All the meshes are rendered by Pyrender\cite{pyrender}.

\end{document}